\documentclass{ectj}

\usepackage{amsfonts,amssymb,graphics,epsfig,verbatim,bm,latexsym,amsmath,url,amsbsy}

\usepackage{xr,float}

\newtheorem{theorem}{Theorem}

\newtheorem{corollary}{Corollary}
\newtheorem{lemma}{Lemma}
\newtheorem{example}{Example}
\newtheorem{remark}{Remark}
\newtheorem{definition}{Definition}

\renewcommand{\thesection}{\arabic{section}}
\renewcommand{\theequation}{\arabic{section}.\arabic{equation}}

\newcommand{\Ep}{{\mathrm{E}}}

\renewcommand{\Pr}{{\mathrm{P}}}

\newcommand{\rem}{\mathrm{rem}}
\newcommand{\divg}{\mathrm{div}}

\received{...}
\accepted{...}
\volume{21}

\title[DML with Riesz Representers]{De-Biased Machine Learning of Global and Local Parameters Using Regularized Riesz Representers}

\author[Chernozhukov, Newey, and Singh]{Victor~Chernozhukov$^{\dagger}$, Whitney~K.~Newey$^{\dagger}$, and
                Rahul~Singh$^{\dagger}$}

\address{$^{\dagger}$MIT Economics, 50 Memorial Drive, Cambridge MA 02142, USA.}
\email{vchern@mit.edu, wnewey@mit.edu, rahul.singh@mit.edu}

\def\AmSTeX{$\cal A$\kern-.1667em\lower.5ex\hbox{$\cal M$}\kern-.125em
    $\cal S$-\TeX}
\def\BibTeX{{\rm B\kern-.05em{\sc i\kern-.025em b}\kern-.08em
    T\kern-.1667em\lower.7ex\hbox{E}\kern-.125emX}}

\begin{document}

    \begin{abstract}
 We provide adaptive inference methods, based on $\ell_1$ regularization, for regular (semi-parametric) and non-regular (nonparametric) linear functionals of the conditional expectation function. Examples of regular functionals include average treatment effects, policy effects, and derivatives. Examples of non-regular functionals include average treatment effects, policy effects, and derivatives conditional on a covariate subvector fixed at a point. We construct a Neyman orthogonal equation for the target parameter that is approximately invariant to small perturbations of the nuisance parameters. To achieve this property, we include the Riesz representer for the functional as an additional nuisance parameter. Our analysis yields weak ``double sparsity robustness'': either the approximation to the regression or the approximation to the representer can be ``completely dense'' as long as the other is sufficiently ``sparse''. Our main results are non-asymptotic and imply asymptotic uniform validity over large classes of models, translating into honest confidence bands for both global and local parameters.
        \keywords{Neyman orthogonality, Gaussian approximation, sparsity}

    \end{abstract}


\section{Introduction}\label{sec:intro}

Many statistical objects of interest can be expressed as a linear functional of a regression function  (or projection, more generally). Examples include global parameters: average treatment effects, policy effects from changing the distribution of or transporting regressors, and average directional derivatives, as well as their local versions defined by taking averages over regions of shrinking volume. This variety of important examples motivates the problem of learning linear functionals of regressions.   Global parameters are typically regular (estimable at $1/\sqrt{n}$ rate), and local parameters are non-regular (estimable at slower than $1/\sqrt{n}$ rates).  Global parameters can also be non-regular under weak identification (for example, in average treatment effects, when propensity scores accumulate mass near zero or one, along a given sequence of models).

Often the regression is high dimensional, depending on many variables such as covariates in a treatment effect model.  Plugging a machine learner into a functional of interest can give a badly biased estimator. To avoid such bias, we use debiased/``double'' machine learning (DML) based on Neyman orthogonal scores that have zero derivative with respect to each first step learner (e.g.,
\cite{neyman1959,belloni2014pivotal,belloni2015uniform,chernozhukov2016locally,chernozhukov2018double,foster2019orthogonal}). Note that the word ``double'' emphasizes the connection to double robustness, a property which orthogonal scores have in this case. Such scores are constructed by adding a bias correction term: the average product of the regression residual with a learner of the functional's Riesz representer (RR). This construction builds upon and is directly inspired by \cite{newey1994asymptotic} where such scores arise in the computation of the semi-parametric efficiency bound for regular functionals.
We also remove overfitting bias (high entropy bias) by using cross-fitting, an efficient form of sample splitting, where we average over data observations different than those used by the nonparametric learners. See, e.g, \cite{schick1986asymptotically} for early use and \cite{chernozhukov2018double} for more recent use in the context of debiased
machine learning.

Using closed-form solutions for Riesz representers in several examples, \cite{chernozhukov2016locally,chernozhukov2018double} defined DML estimators in high dimensional settings and established their good properties. In comparison, the new approach proposed in this paper has the following advantages and some limitations:
    \begin{enumerate}
       \item  We provide a novel algorithm based on $\ell_1$ regularization to \textit{automatically estimate} the Riesz representer from the empirical analog of equations that implicitly characterize it. 
       \item  Even when a closed-form solution for the Riesz representer is available, the method avoids estimating each of its components. For example, the method avoids explicit density derivative estimation for the average derivative, and it avoids inverting estimated propensity scores for average treatment effects. 
       \item  The adaptive inference theory covers both regular objects (estimable at the $1/\sqrt{n}$ rate)  and nonregular ones (with rates $L/\sqrt{n}$, where
       $L \to \infty$ is the operator norm of the  linear functional). 
       \item As far as we know, the adaptive inference theory given here is the first non-asymptotic Gaussian approximation analysis of de-biased machine learning. 
    \item  Our approach remains interpretable under misspecification, estimating a linear functional of the projection rather than regression. (This point is made explicit in Section~\ref{sec:estimation}).
    \item  We provide a non-asymptotic analysis when using the $\ell_1$-penalized method to learn the regression, and an asymptotic analysis when using other modern machine learning estimators to learn the regression.
    \item  The current analysis focuses on linear functionals. In follow-up work, \citet*{chernozhukov2018learning} extend the approach to nonlinear functionals through a linearization.
    \end{enumerate}
This paper is a revised version of \citet*{chernozhukov2018robins} that gave an algorithm based on $\ell_1$ regularization for automatically estimating the Riesz representer. This version is distinguished from \citet*{chernozhukov2018robins}, \cite{chernozhukov2018double}, \cite{chernozhukov2016locally}, and \citet*{chernozhukov2018learning} in covering local objects that are estimated at a rate slower than $1/\sqrt{n}$. Providing debiased machine learning for such local objects is an important contribution of this paper. 

Sections~\ref{sec:target} and~\ref{sec:applications} present the main ideas for a general audience. In Section~\ref{sec:target}, we define global, local, and perfectly localized linear functionals of the regression, and provide orthogonal representations for these functionals. In Section~\ref{sec:applications}, we present two empirical examples: local and global average treatment effects, and local and global average derivatives. 

Sections~\ref{sec:estimation} and~\ref{sec:general} are theoretical. In Section~\ref{sec:estimation}, we provide estimation theory, demonstrating concentration and approximate Gaussianity of the DML estimator with regression and Riesz representer estimated via regularized moment conditions. We provide rates of convergence for estimating the Riesz representer, giving both fast rates under approximate sparsity and slow rates under the dense model. In Section~\ref{sec:general}, we demonstrate asymptotic consistency and Gaussianity of the DML estimator with regression estimated via general machine learning.

The supplement provides supporting material. In Section~\ref{sec:related}, we give a detailed account of how our work relates to previous and contemporary work. In Section~\ref{sec:preliminary}, we review prelimaries of functional analysis.  In Section~\ref{sec:examples}, we analyze the structure of the leading examples, providing bounds on operator norm, variance of the score, and kurtosis. Finally, we provide proofs for each section.

\section{Overview of target functionals, orthogonal representation, estimation, and inference}\label{sec:target}

\subsection{Target functionals} 
We consider a random element $W$ with distribution $P$  taking values $w$ in its support $\mathcal{W}$. Denote the $L^q(P)$ norm of a measurable function $f: \mathcal{W} \to \Bbb{R}$ and also the $L^q(P)$ norm of random variable $f(W)$ by $\| f \|_{P,q} = \| f(W)\|_{P,q}$.  For a differentiable map $x \mapsto g(x)$, from $\mathbb{R}^d$ to $\mathbb{R}^k$,  we use $\partial_{x'} g$ to abbreviate the partial derivatives  $(\partial/\partial x') g(x)$, and we use $\partial_{x'} g(x_0)$ to mean $\partial_{x'} g (x) \mid_{x = x_0}$, etc.  We use $x'$ to denote the transpose of a column vector $x$.

Let $(Y,X)$ denote a random sub-vector of $W$ taking
values  in their support sets, $y \in \mathcal{Y} \subset \Bbb{R}$ and $x \in \mathcal{X} \subset \Bbb{R}^{d_x}$, where $d_x=\infty$ is allowed.  Let $F$ denote the law of $X$. We define $$x \mapsto \gamma_0^\star(x) := \Ep [Y\mid X=x],$$ as the unknown regression function of $Y$ on $X$. We  consider the convex parameter space $\Gamma_0$ for $\gamma_0^\star$ with 
elements $\gamma$.  (Later, in the theoretical sections,  we generalize and replace the regression function by a projection).

Our goal is to construct high-quality inference methods for real-valued linear functionals of $\gamma_0^\star$. To present examples below we need to endow $\gamma_0^\star$ with a causal interpretation, which requires us to assume that it is  a structural function, invariant to the changes in the distribution of $X$ under policies described below. This property is not guaranteed for an arbitrary regression problem. For the reader who is unfamiliar with these concepts, we note that a simple sufficient condition for invariance is follows: given a stochastic process $x \mapsto Y(x)$, called potential outcomes or structural function, vector $X$ is generated to follow distribution $F$ independently of  $x \mapsto Y(x)$ and $Y$ is generated as $Y = Y(X)$. In this  case we  have $\gamma_0^\star(x) = \Ep Y(x)$ for any $F$.  This condition is conventionally called exogeneity in econometrics and random assignment in statistics. The measurability requirement here is that $(x, \omega)\mapsto Y(x,\omega)$ is a measurable map. We refer to \cite{imbens2015causal}, \cite{hernan2019causal}, and \cite{peters2017elements} for the relevant formalizations that enable causal interpretation.
 
\begin{example}[Average treatment effect]\label{ex:ate}    Let
$X = (D,Z)$ and $\gamma_0^\star(X) = \gamma_0^\star(D,Z)$, where $D \in \{0,1\}$ is the indicator of the receipt of the treatment.
Define $$
\theta^\star_0 = \int (\gamma_0^\star(1,z) - \gamma_0^\star(0,z))  \ell(x) d F(x),$$
where $x	 \mapsto \ell(x)$ is a weighting function. This statistical parameter is a weighted average treatment effect under the standard conditional exogeneity assumption, which guarantees that  $\gamma_0^\star$ is invariant to changes in the
distributions of $D$ conditional on $Z$. The assumption requires $D$ to be independent of the potential outcome process $d \mapsto Y(d,Z)$ and  outcome to be generated as $Y= Y(D,Z)$, so that $\gamma_0^\star(d,z) = \Ep [Y(d,Z) \mid Z=z]$. Here $\gamma_0^\star$ is invariant to changes in the conditional distributions of $D$, but not to the changes in the distribution of $Z$. 
\end{example} 

Here and below, a weighting function is a measurable function $x \mapsto \ell(x) $ such that $\int \ell dF = 1$
and $\int \ell^2 dF < \infty$. In this example, setting 
\begin{itemize}
\item $\ell(x)  = 1$ gives average treatment effect (ATE) in the entire population, 
\item $\ell(x)= 1(d=1)/P(D=1)$ gives the ATE for the \textit{treated} population, 
\item $\ell(x) = 1( z \in N)/P(Z \in N)$ the ATE
conditional on  $Z$ in the neighborhood $N$,
\end{itemize}
and so on. We can model small neighborhoods $N$ as shrinking in volume with the sample size. The local weighting
and kernel weighting discussed below are applicable to all key examples.  Moreover, they are combinable with other weighting functions so that, for example, we can target inference on local average treatment effects for the treated.

\begin{example}[Policy effect from changing distribution of $X$]\label{ex:policy1}  The average causal effect  of the policy that shifts the distribution of covariates from $F_0$ to $F_1$ with the support contained in $\mathcal{X}$, when $\gamma_0^\star$ is invariant over $\{F, F_0, F_1\}$, for the weighting function $x \mapsto \ell(x)$, is given by:
$$
\theta^\star_0 =  \int \gamma_0^\star(x) \ell(x) d G( x); \quad G(x) = F_1( x) - F_0(x). 
$$
Exogeneity is a sufficient condition for the stated invariance of $\gamma_0^\star$.
\end{example} 

\begin{example}[Policy effect from transporting $X$]\label{ex:policy2} A weighted average  effect of changing covariates $X$ according to  a transport map $X \mapsto T(X)$, where $T$ is deterministic measurable map from $\mathcal{X}$ to $\mathcal{X}$, with the weighting function $x \mapsto \ell(x)$, is given by:
$$
\theta^\star_0 = \int[ \gamma_0^\star(T(x)) - \gamma_0^\star(x) ]\ell(x) d F(x). 
$$
This has a causal interpretation if the policy induces the \textit{equivariant} change in the regression function, namely  the outcome $\tilde Y$ under the policy obeys
$\Ep [\tilde Y |X] = \gamma_0^\star(T(X))$. Exogeneity is a sufficient condition. \end{example} 

\begin{example}[Average directional derivative]\label{ex:deriv}  In the same settings as the previous example,  a weighted average derivative of  a continuously differentiable $\gamma_0$ with respect to component vector $d$ in the direction $d \mapsto t(x)$ and weighed by $x \mapsto \ell(x)$ is  the linear functional of the form:
$$
\theta^\star_0 = \int \ell(x) t(x) '\partial_d \gamma_0^\star(d,z) dF(x).
$$
In causal analysis, $\theta^\star_0$ is an approximation to  $1/r$ times the  average causal effect  of the policy that shifts
the distribution of covariates via the map $X=(D,Z) \mapsto T(X) = (D +  r t(X), Z)$ for small $r$,  weighted by $\ell(X)$.
Here we require that $(d,x) \mapsto \partial_d \gamma_0^\star(x)$ exists and is continuous on $\mathcal{X}$.
\end{example}

In this example, consider the case when $X=(D,Z)$ consists of continuous treatment variable $D$ and
covariates $Z$. Further suppose $\ell(x)=\ell(d)$ and $t(x)=1$. Then the parameter of interest
is $\theta_{0}^{\ast}=\Ep[\ell(D)T(D)],$ where $T(d)=\Ep[\partial_d\gamma_{0}^{\ast
}(D,Z)|D=d].$ When $Y=Y(D)$ for a potential outcome process $Y(d)$
that is independent of treatment $D$ conditional on the covariates $Z$ and
differentiable in $d,$ it was shown by \cite{altonji2005cross} and \cite{florens2008identification} that $T(d)=\Ep[\partial_d Y(D)|D=d],$ which is an average
treatment effect on the treated. Thus $\theta_{0}^{\ast}$ is a weighted
average of the effect of treatment on the treated and would be equal to $T(d)$
for the perfectly localized $\ell(d)=1(D=d)/f_{D}(d),$ where $f_{D}(d)$ is the
pdf of $D.$ Also for $\ell(d)\equiv1$, \cite{imbens2009identification} showed that
$\theta_{0}^{\ast}=\Ep[T(D)]=\Ep[\partial_d Y(D)]$, which is an average
treatment effect. See also \cite{rothenhausler2019incremental}.

In Example~\ref{ex:deriv}, we consider the case where the variable of differentiation is also the variable of localization. As explained above, this case corresponds to effects of continuous treatments, and it turns out to require extra care in Section~\ref{sec:examples}. The other possible case is where the variable of differentiation is different than the variable of localization. Such a case turns out to be simpler and is handled by similar arguments as Examples~\ref{ex:ate},~\ref{ex:policy1}, and~\ref{ex:policy2} in Section~\ref{sec:examples}.

All of these statistical parameters play an important role in causal inference, counterfactual  decompositions, and predictive analyses. Introduction of the weighting function $\ell(X)$ allows us to study subgroup effects and local effects, and these will be covered by our non-asymptotic results and asymptotic results. All of the above examples can be viewed as real-valued linear functionals of the regression function.

\begin{definition}[Target parameter] Our target is the real-valued linear functional of $\gamma_0^\star$:
\begin{equation} \label{eq: def theta}
\theta^\star_0= \theta(\gamma_0^\star), \text{ 
where } \gamma \mapsto \theta (\gamma) := \Ep m(W, \gamma),\end{equation}
 $\gamma \mapsto m(w, \gamma)$ is a linear operator for each $w \in \mathcal{W}$, defined on $\Gamma= \mathrm{span}(\Gamma_0)$, and the  map $w \mapsto m(w, \gamma)$ is measurable with finite second moment under $P$ for each $\gamma \in \Gamma$.
 \end{definition}

The linear operator $\gamma \mapsto \theta(\gamma)$ has the following generating function $m$ in these examples:\begin{enumerate}
\item[2.1]  $m(w, \gamma)  = (\gamma(1,z) - \gamma(0,z)) \ell(x) $;  
\item[2.2] $m (w, \gamma) = m(\gamma) = \int \gamma(x) \ell(x) d G(x); \ \ G(x) = F_1(x) - F_0(x)$;
\item[2.3]  $m(w, \gamma) = \ell(x)(\gamma(T(x)) - \gamma(x))$;
\item[2.4]  $m(w, \gamma) =  \ell(x) t(x)'\partial_d \gamma(x)$.
 \end{enumerate}
In these examples, we can recognize the dependency on the weighting function by writing
$
m(w, \gamma; \ell).
$
In Examples~\ref{ex:ate},~\ref{ex:policy2}, and~\ref{ex:deriv} we can decompose
$m(w, \gamma;\ell) = m_0(w, \gamma)\ell(x).$

Estimation of some parameters of the form in Definition 2.1 is very straightforward, such as $\Ep[w(X)\gamma_0(X)]$ for a known function $w(x)$. These can be estimated as the sample mean of $w(X)Y$. Such simple estimation is not possible for the causal, counterfactual parameters in Examples~\ref{ex:ate},~\ref{ex:policy1},~\ref{ex:policy2}, and~\ref{ex:deriv}. The approach of this paper provides estimators for these counterfactual parameters and can be used for many others.

Our local functionals are defined by using the weight function that  localizes the functionals around value $d_0$ of a low-dimensional vector component $D$.  Here $D$ is a $p_1$-dimensional component of vector $X$.

We consider the weighting function
\begin{equation}
\ell_h(D) = \frac{1}{h^{p_1}} K\left( \frac{d_0-D}{h}\right)/\omega, \quad \omega = \Ep \left[ \frac{1}{h^{p_1}} K\left( \frac{d_0-D}{h}\right) \right],  \quad h \in \Bbb{R}_+,
\end{equation}
where $K: \Bbb{R}^{p_1} \to \Bbb{R}$ is a kernel function of order $\mathsf{o}$ such that $\int K = 1$ and 
$$\int (\otimes^m u)  K(u) du = 0,  \quad \text{ for } m=1,..., \mathsf{o}-1,$$ with its support contained in the cube $[-1,1]^{p_1}$. The simplest example is the box kernel with $K(u) = \times_{j=1}^{p_1}1(-1 < u_j < 1)/2$,
which is of order $\mathsf{o} = 2$.

\begin{remark} For the case of Example 1, the localization is understood to be with respect to any continuous component $\tilde D$ of covariate vector $Z$, but of course not the binary treatment $D$.
\end{remark}

To present the main results in the most clear way, we assume that $\ell_h$ is known,
i.e. $\omega$ is known. Our main results also hold for one sided kernels. We leave to future work the application of this theory to settings with one sided limits, e.g. regression discontinuity design.  

\begin{definition}[Local and localized functionals]
 We consider the local functional
$$
 \theta(\gamma_0^\star; \ell_h) := \Ep m(W, \gamma_0^\star; \ell_h),
$$
as well as the (perfectly) localized functional
$$
\theta(\gamma_0^\star; \ell_0) := \lim_{h \to 0} \theta(\gamma_0^\star; \ell_h).
$$
\end{definition}

The difficulty in targeting
localized functionals is that they are not pathwise differentiable. A key quantity in the analysis is the operator norm (the modulus of continuity) of $\gamma \mapsto \theta (\gamma)$ on $\Gamma$, defined as
\begin{equation} L:= \sup_{\gamma \in \Gamma\setminus\{0\}}  |\theta(\gamma)|/\| \gamma\|_{P,2}.
\end{equation}
We consider $L=\infty$ as non-regular cases, e.g. perfectly-localized functionals. We also consider cases
where $L \to \infty$ as $n \to \infty$ as non-regular.  Indeed, the latter case arises from approximating the functional
with $L=\infty$ by functionals where $L \to \infty$, e.g. local functionals with $h \to 0$.  
The $L=\infty$  case also arises  in triangular array asymptotics where $P$ changes with $n$. 
The asymptotic thought experiment, where $L \to \infty$,  approximates
non-asymptotic cases where $L$ is high. We emphasize that we derive both non-asymptotic
results and their asymptotic corollaries (which lead to simplified statements conveying key qualitative features of 
non-asymptotic  results).

 \subsection{Building an orthogonal representation of the target functional}

Equation~(\ref{eq: def theta}) can be thought of as a direct formulation of the target parameter. Next we introduce a dual formulation and finally an orthogonal formulation. Towards this end, we define the Riesz representer $\alpha_0$.


\begin{definition}[Linear and minimal linear representer]  A linear representer (also called a Riesz representer) for the linear functional  $\gamma \mapsto \theta(\gamma)$ is $\alpha_0 \in L^2(F)$ such that 
\begin{equation}\label{eq: Riesz}
 \theta(\gamma) =\Ep \gamma(X) \alpha_0(X), \text{ for all } \gamma \in  \Gamma.  
  \end{equation}
If $\alpha_0  \in \bar \Gamma := \text{closure}(\Gamma)$ in $L^2(F)$, we call it the minimal representer and denote it by $\alpha^\star_0$;  if not, we call it  a representer. Any representer can be reduced to the minimal representer by projecting it onto $\bar \Gamma$. 
 \end{definition}
 
 A minimal linear representer exists if and only if $L< \infty$, as a consequence of the Riesz--Frechet theorem; see Lemma~\ref{lemma: ERR} below.   Therefore, when $L< \infty$, we define the following dual linear representation for the target parameter\begin{equation}\label{RRR}
 \theta^\star_0 = \theta(\alpha_0^\star);  \ \ \theta(\alpha) := \Ep [\alpha(X) Y]. \end{equation}

 To motivate the upcoming orthogonal representation, we note that either the direct or the dual identification strategies can be used for direct plug-in estimation, but this does not give good estimators, as explained in the following technical remark.
 
\begin{remark}[Non-orthogonality of direct and dual formulations]Even if we knew expectation operator $\Ep$  and use $\theta(\hat \gamma)$ or $\theta(\hat \alpha)$ as the estimator for $\theta^\star_0$, this estimator would
 have high biases. Indeed, neither $\gamma \mapsto \theta(\gamma)$ nor $\alpha \mapsto \theta(\alpha)$ are
  orthogonal to local perturbations $h \in \Gamma$ of $\gamma_0^\star$ or $\bar h \in \Gamma$ of $\alpha_0^\star$, namely $$
\partial_t \theta(\gamma_0^\star + t h)   \Big |_{t =0}  = \Ep m (W, h) \neq 0, \ \ 
\partial_t \theta(\alpha_0^\star + t  \bar h )   \Big |_{t =0}  =\Ep \gamma_0^\star (X) \bar h(X) \neq 0.
$$
Consequently, the quantities $\Ep m (W, \hat \gamma - \gamma_0^\star)$  and $ \Ep \gamma_0^\star (\hat \alpha - \alpha_0^\star)$ 
are first order biases for $\theta(\hat \gamma)$ and $\theta(\hat \alpha)$. The regularized estimators $\hat \gamma$ or $\hat \alpha$ exploit structure of $\gamma_0^\star$
and $\alpha_0^\star$  to estimate them well in high-dimensional problems, but they exhibit biases that vanish at rates slower than $1/\sqrt{n}$, which makes  $\theta(\hat \gamma)$ and $\theta(\hat \alpha)$  converge at the same slow rate.
\end{remark}


 Therefore we proceed to construct another representation for $\theta^\star_0 $ that has the required Neyman orthogonality structure.

 \begin{definition}[Orthogonal representation for the target functional]
 We have \begin{equation}\label{DR}
 \theta^\star_0 = \theta(\alpha_0^\star,\gamma_0^\star);  \ \ \theta(\alpha, \gamma) := \Ep [m(W, \gamma) + \alpha(X) (Y - \gamma(X))],
 \end{equation}
where  $(\alpha, \gamma)$ are the nuisance parameters with the true value $(\alpha_0^\star,\gamma_0^\star)$.
  \end{definition}
 
Unlike the direct or dual representations for the functional, this representation is Neyman orthogonal to perturbations $(\bar h, h)$ $\in \Gamma^2$ of  $(\alpha_0^\star,\gamma_0^\star)$ such that
\begin{equation}\label{eq:invariance}
\frac{\partial}{\partial t}  \theta(\alpha_0^\star + t \bar h, \gamma_0^\star + t h)  \Big |_{t=0} = \Ep m(W, h) - \Ep \alpha_0^\star(X) h(X) + \Ep [(Y - \gamma_0^\star(X)) \bar h(X)] = 0.
\end{equation}
In fact, a stronger property holds 
\begin{equation}
 \theta(\alpha, \gamma) - \theta(\alpha^\star_0, \gamma^\star_0) = - \int (\gamma - \gamma^\star_0) ( \alpha - \alpha_0^\star) dF,
\end{equation}
which implies (\ref{eq:invariance}) as well as double robustness. 

The Neyman orthogonality property states that the representation of the target parameter $\theta_0$ in terms
of the nuisance parameters $(\alpha, \gamma)$ is invariant to the local perturbations of the values of the nuisance parameter.
This property makes the orthogonal representation  an excellent basis for constructing high quality point and interval estimators of $\theta^\star_0$ in modern high-dimensional settings when we will be plugging-in biased estimators in lieu of  $\gamma_0^\star$ and $\alpha_0^\star$, where the bias occurs because of the regularization (see, e.g., \cite{chernozhukov2016locally,chernozhukov2018double}).   

Both $\gamma_0^*$ and $\alpha_0^*$ are identified, $\gamma_0^*$ as $\Ep[Y|X]$ and $\alpha_0^*$ by virtue of the consistent estimator we give in Section~\ref{sec:preview}. Identification allows us to use the orthogonal representation to estimate target parameters.

\subsection{The case of finite-dimensional linear regression}\label{sec:linear}  It is instructive to consider the case of linear finite-dimensional regression. Consider $x \mapsto b(x) = \{ b_j(x)\}_{j=1}^p$ as a $p$-dimensional dictionary of basis
functions with $b_j \in L^2(F)$ for each $j=1,...,p$.   The regression function is assumed to obey the linear functional form $\gamma^\star_0 = b'\beta_0$ for some $\beta_0$.  Also define
$$
G= \Ep b(X)b(X)', \quad M = \Ep m(W, b).
$$

First, observe that for $\gamma = b'\beta$, 
$$\theta (\gamma) = \Ep m(W, b'\beta) =   \Ep m(W, b)' \beta = M \beta.$$  For instance, in Examples~\ref{ex:ate},~\ref{ex:policy1},~\ref{ex:policy2}, and~\ref{ex:deriv}:
\begin{enumerate} 
\item[] 2.1 $M =  \Ep (b(1,Z) - b(0,Z))\ell(X)$\quad 
2.2 $M= \int b \ell (d F_1- dF_0)$,
\item[] 2.3 $M= \Ep (b(T(X)) -  b(X) )\ell(X)$\quad 
2.4 $M = \Ep \partial_d b(D,Z) t(X) \ell(X)$.
\end{enumerate}

Second, we make a guess that the linear representer $\alpha^\star_0$ to be of the form
$
\alpha^\star_0(x)= b(x)' \rho_0,
$
for $\rho_0$ defined below. We can define the parameters
$\beta_0$ and $\rho_0$ as any minimal $\ell_1$-norm solutions to the system of equations:
\begin{equation}\label{eq: linear}
\min \|\beta\|_1 + \| \rho\|_1: \quad G \beta = \Ep Y b(X) , \quad   G \rho = M.
\end{equation}
In particular, if $G$ is full rank, the solutions are 
 $\beta_0= G^{-1} \Ep b(X) Y$
and $\rho_0 = G^{-1} M$.

We now verify the representation property for our guess:
$$
\Ep \gamma(X) \alpha^\star_0(X) = \Ep \beta' b(X)b(X)'\rho_0  = \beta' G \rho_0 = \beta'M =   \theta(\gamma),
$$
for all $\beta$'s and hence all $\gamma$'s. The operator norm of $\theta (\gamma) = M'\beta$ is given by
$$
L= \sup_{\beta \in \Bbb{R}^p\setminus \{0\}}\frac{ | M'\beta| }{\sqrt{\beta'G\beta}} =
\sup_{\beta \in \Bbb{R}^p\setminus \{0\}} \frac{|\beta'G\rho_0|}{\sqrt{\beta'G\beta}} = \sqrt{\rho_0' G \rho_0} < \infty.
$$

We conclude that direct, dual, and orthogonal representations are given by
$$
\theta(\gamma) = M'\beta;  \quad \theta(\alpha) = \rho' \Ep  b(X)Y ; \quad \theta(\gamma, \alpha) = 
M'\beta +  \rho' \Ep  b(X)Y -   \rho' G\beta,
$$
where $\beta$ is  $\gamma$'s parameter and $\rho$ is $\alpha$'s parameter.  These representations appear to be both novel
and useful.

 \subsection{The case of infinite-dimensional regression} In the infinite-dimensional case, we can employ the Riesz--Frechet representation theorem and Hahn--Banach extension theorem to establish existence of the linear  Riesz representer. 

 \begin{lemma}[Extended Riesz representation]\label{lemma: ERR}  (i) If $L< \infty$, there exists a unique minimal
 representer $\alpha_0^\star \in \bar \Gamma$ and $L= \| \alpha^{\star }_0\|_{P,2}$. (ii) If there exists a linear representer $\alpha_0$ on $\Gamma$ with $\|\alpha^{}_0\|_{P,2} < \infty$, then  $L = \|\alpha^{\star }_0\|_{P,2}  \leq \|\alpha^{}_0\|_{P,2}  < \infty$, where
 $\alpha^{\star}_0$, obtained by projecting $\alpha_0$ onto $\bar \Gamma$, is the unique minimal representer.   In both cases
 $\gamma \mapsto \theta(\gamma)$ can be extended to $\bar \Gamma$ or to the entire $L^2(F)$ with the modulus of continuity $L$.
\end{lemma}

The first part of the lemma shows (implicit) existence of a linear representer when $L < \infty$. Our estimation results will rely \textit{only on the existence} of minimal representers.  In some cases, however, we may utilize the closed-form solutions for linear representers (see, e.g., Section~\ref{sec:examples} for the key examples),  to improve the basis functions for estimating the minimal representers.   There is also an efficiency reason to work with minimal representers rather than any linear representer, as highlighted in Section~\ref{sec:estimation} analyzing semi-parametric efficiency.

 \subsection{Informal preview  of estimation and inference results}\label{sec:preview}

 Our estimation and inference will exploit empirical analogs of both 
 the orthogonal representation of the parameter (\ref{DR}) and the equation defining the RR property (\ref{eq: Riesz}). 
 
 To approximate the regression function and the RR, we consider the $p$-vector of dictionary functions $b$, where the dimension $p$ of the dictionary can be large, potentially much larger than $n$. We approximate $\alpha^\star_0$ by a linear form $b'\rho_0$, and we approximate
$\gamma^\star_0$ by a linear form $b'\beta_0$, and estimate the parameters using the algorithms below.

\begin{enumerate}
    \item Let $(W_i)_{i=1}^n= (Y_i, X_i)_{i=1}^n$ denote i.i.d. copies of data vector $W$.  We use cross-fitting to avoid biases from overfitting that can arise in high-dimensional settings.  To this end, let $(I_{1},..., I_K)$ be a partition of the
observation index set $\{1,...,n\}$ into $K$ distinct subsets of about equal
size. Let $\mathbb{E}_A f = \mathbb{E}_A f(W)$ denote the empirical average of $f(W)$ over $i \in A \subset \{1,..., n\}$: $\mathbb{E}_{A} f  := \mathbb{E}_{A} f(W) = | A | ^{-1} \sum_{i \in A} f(W_i).$ 
    \item For each block $k=1,...,K$,  we obtain generalized Dantzig selector (GDS) estimates $\hat \alpha_k= b'\hat \rho_k$ and $\hat \gamma_k= b'\hat\beta_k$, where
  \begin{equation}
\begin{array}{l} \hat \rho_k = \arg \min_{\rho \in \Bbb{R}^p} \| \rho\|_1  : \|  \hat{D}^{-1}\left\{\mathbb{E}_{I^c_k}  m(W, b)  -  \mathbb{E}_{I^c_k} b(X) b(X)'\rho\right\} \|_\infty \leq \lambda_\rho,\\
 \hat \beta_k = \arg \min_{\beta \in \Bbb{R}^p}  \| \beta \|_1  : \| \hat{D}^{-1}\left\{\mathbb{E}_{I^c_k}  (Y- b(X)'\beta) b(X))\right\}\|_\infty \leq \lambda_\beta,
 \end{array}
 \end{equation}
  where $I^c_k = \{1,...,n\}\setminus I_k$ is the set of observation indices leaving $I_k$ out, $\lambda$'s are tuning parameters, and $\hat{D}$ is a scaling detailed in Section~\ref{sec:diff}. Typically $\lambda$'s scale like $\sqrt{ \log (p \vee n)/n}$; Section~\ref{sec:applications}  provides concrete choices. 
    \item The DML estimator is an average of estimated orthogonal representations over $k$:
\begin{equation}
\hat{\theta}=\frac{1}{n}\sum_{k=1}^{K}\sum_{i\in I_{k}}\{m(W_{i}%
,\hat{\gamma}_{k})+\hat{\alpha}_{k}(X_{i})[Y_{i}-\hat{\gamma}_{k
}(X_{i})]\}. \label{Estimator}%
\end{equation}
The estimator of its asymptotic variance is
\begin{equation}
\hat{\sigma}^2=\frac{1}{n}\sum_{k=1}^{K}\sum_{i\in I_{k}}\{m(W_{i}%
,\hat{\gamma}_{k})+\hat{\alpha}_{k}(X_{i})[Y_{i}-\hat{\gamma}_{k
}(X_{i})]-\hat{\theta}\}^2. \label{Estimator_var}%
\end{equation}
\end{enumerate}
We remark that the RR estimator in step 2 is of Dantzig selector type, but is not exactly the Dantzig selector, requiring some new analysis. We use the GDS rather than series or spline estimation to accommodate high dimensional specifications for the regression and Riesz representer.
 
 The dictionary $b(x)$ is very important for the GDS estimator. This dictionary should be chosen so that linear combinations of $b(x)$ can approximate in mean square any element of $\Gamma$. For example if $\Gamma$ is the set of linear combinations of an infinite sequence of regressors, as for a high dimensional regression, then $b(x)$ could be chosen as the first $p$ elements of that sequence. Also $p$ can be chosen flexibly, because $p$ will be allowed to grow faster than the sample size, as specified in the asymptotic theory to follow. In practice multiple choices of $p$ could be tried.
 
 Next, we state the key concentration and approximate Gaussianity results informally. Key quantities in the analysis are the ``true'' score  and its moments:
 $$
\psi_{0}^\star(W) :=  \theta_0^\star-m(W, \gamma_0^\star)-\alpha^\star _0(X)(Y- \gamma_0^\star(X)),\ \ \sigma^2 := \Ep \psi^2_{0}(W), \ \ \kappa^3 := \Ep |\psi^3_{0}(W)|.
 $$

We establish conditions under which 
\begin{equation}\label{eq:SC-1}
\| \hat \gamma_k - \gamma_0^\star\|_{P,2} + \| \hat \alpha_k - \alpha_0^\star\|_{P,2}/\sigma \to 0,\quad 
\sqrt{n}\int (\hat \gamma_k - \gamma^\star_0) (\hat \alpha_k - \alpha^\star_0) dF/\sigma  \to 0.
\end{equation}
These include a bound on the $\ell_1$ norm of coefficients and that either the regression function or 
the RR is approximately sparse with the effective dimension $s$ less than $\sqrt{n}$. This allows either nuisance parameter to be completely dense.
 
Given that (\ref{eq:SC-1}) holds, we establish  that the resulting de-biased (or ``double'') machine learning (DML) estimator $\hat \theta$ is \textit{adaptive}, namely it is approximated up to the error $o(\sigma/\sqrt{n})$ by the oracle estimator
$$
\bar \theta := \theta^\star_0 - n^{-1} \sum_{i =1}^n \psi_0(W_i),
$$
where the \textit{oracle} knows the scores $\psi_0$.  Hence the approximate deviation of $\hat \theta$ from $\theta_0^\star$ is determined by $\|\psi_{0}\|_{P,2}/\sqrt{n}$, which is the standard deviation of the oracle estimator.

Consequently, $\hat \theta$ concentrates in a $\sigma/\sqrt{n}$ neighborhood of the target  with deviations controlled by the normal laws, 
$$
\sup_{t \in \Bbb{R} }\Big |\Pr ( \sqrt{n}(\hat \theta - \theta^\star_0)/ \sigma  \leq t) -  \Pr( N(0,1) \leq t) \Big |  \leq A (\kappa/\sigma)^3/\sqrt{n} + \mathrm{error}_n \to 0,
$$
where the $\mathrm{error}_n$ bound is non-asymptotic and tends to zero as $n \to \infty$. Of course, $\sigma/ \sqrt{n} \to 0$ is required
for concentration. The non-asymptotic bound automatically implies the uniform validity of results over large classes of probability laws $P$ for $W$.   

There are two cases to consider:
\begin{enumerate}
 \item\textsc{Regular case}: \textit{the parameters $\sigma$, $\kappa/\sigma$, and $L$ are bounded, leading to $1/\sqrt{n}$ concentration, adaptation, and Gaussian approximation.}
 \item \textsc{Non-regular case}, \textit{the parameters $\sigma, \kappa/\sigma$, and $L$ diverge, so that we need 
$$
\sigma/\sqrt{n} \to 0, L/\sqrt{n} \to 0, (\kappa/\sigma)/\sqrt{n}  \to 0,
$$
for $\sigma/\sqrt{n}$ concentration, adaptation, and Gaussian approximation.}
\end{enumerate}

As we show in Section~\ref{sec:examples}, in the case of local functionals, the latter condition can be more succinctly stated as
$$
(\kappa/\sigma) \lesssim \sigma \asymp L ,  \quad L/\sqrt{n}  \to 0.
$$

Finally, we establish that we can \textit{transfer} learning and inference guarantees for local functionals to 
those for the \textit{(perfectly) localized} functionals if the localization bias is sufficiently small, namely
$$
\sqrt{n} (\theta(\gamma_0^\star; \ell_h) - \theta(\gamma_0^\star; \ell_0)/\sigma \to 0.
$$

We think it is remarkable that a single inference theory covers both regular and non-regular cases,
and provides uniform validity over large classes of $P$.

\section{Applications}\label{sec:applications}

\subsection{Global and local effects of 401(k) eligibility on net financial assets}

First, we use our method to answer a question in household finance: what is the average treatment effect of 401(k) eligibility on net financial assets (over a horizon of about two years)? We follow the identification strategy of \cite{poterba1994} and \cite{poterba1995}, who assume selection on observables. The authors assume that when 401(k) was introduced, workers ignored whether a given job offered 401(k) and instead made employment decisions based on income and other observable job characteristics; after conditioning on income and job characteristics, 401(k) eligibility was exogenous at the time. This empirical question corresponds to Example~\ref{ex:ate}.

We use data from the 1991 US Survey of Income and Program Participation (\cite{chernozhukov2018double_data}), using sample selection and variable construction as in \cite{abadie2003semiparametric} and \cite{chernozhukov2004effects}. The outcome $Y$ is net financial assets defined as the sum of IRA balances, 401(k) balances, checking accounts, US saving bonds, other interest-earning accounts, stocks, mutual funds, and other interest-earning assets minus non-mortgage debt. The treatment $D$ is an indicator of eligibility to enroll in a 401(k) plan. The raw covariates $X$ are age, income, years of education, family size, marital status, two-earner status, benefit pension status, IRA participation, and home-ownership. We impose common support of the propensity score for the treated and untreated groups based on these covariates, yielding $n=9869$ observations. We consider the fully-interacted specification $b(D,X)$ of \cite{chernozhukov2018double} with $p=277$ including polynomials of continuous covariates, interactions among all covariates, and interactions between covariates and treatment status.

Tables~\ref{table_ATE} and~\ref{table_ATE2} summarize results for the entire population and for each quintile of the income distribution. We use $K=5$ folds in cross-fitting. To estimate the RR, we use the generalized Dantzig selector (GDS) procedure introduced in the present work.  To estimate the regression, we use GDS, Lasso, random forest, or neural network. GDS is implemented using the tuning procedure described in Section~\ref{sec:diff}. Lasso is implemented using the tuning procedure described in \citet*{chernozhukov2018learning}. Random forest and neural network are implemented with the same settings as \cite{chernozhukov2018double}, i.e. with 1000 trees or a single hidden layer of eight neurons, respectively. We find ATE of $7608$ $(1395)$ using GDS for both the RR and the regression. This ATE estimate is stable across different choices of regression estimator. We find that localized ATE is not statistically significant for the second quintile, and it is statistically significant, positive, and strongly heterogeneous for the other quintiles. Interpreting the relatively high effect of 401(k) eligibility for the first quintile is a question for future research.

For comparison, \cite{chernozhukov2018double} report ATE of $7170$ $(1398)$ by DML, which estimates the RR by estimating the propensity score and plugging it into the RR functional form. Though these two estimators are asymptotically equivalent under correct specification, our estimator avoids the estimated propensity score in the denominator which could cause numerical instability. The ATE results are broadly consistent with \cite{poterba1994} and \cite{poterba1995}, who use a simpler specification motivated by economic reasoning.  The localized ATE estimates by income quintile group appear to be new empirical results and are of interest in their own right. In Section~\ref{sec:diff} we report analogous estimates without debiasing. Without debiasing, the GDS and Lasso estimates of ATE are attenuated due to regularization. The bias is smaller for the random forest and neural network estimates of ATE.

\begin{table}[h]
    \centering
    \caption{Average treatment effect of 401(k) eligibility on net financial assets. Localized average
    treatment effects are reported by income quintile groups. The regression is estimated by GDS or Lasso. Standard errors are reported in parentheses.}
    \begin{tabular}{@{}ccccccc@{}}
    \hline 
        Income quintile & N treated & N untreated & \multicolumn{2}{c}{GDS}  &\multicolumn{2}{c}{Lasso}   \\
        \hline 
All & 3682 & 6187   &   7607.95   &   (1394.92)& 7733.31   &   (1416.46)\\
1 & 272 & 1702   &   4500.33   &   (924.12)&4477.43  &   (920.31)\\
2 & 527 & 1447   &    1051.60   &   (1501.03)&  1119.06   &   (1500.78)\\
3 & 755 & 1219   &    5204.93   &   (1199.87)&4919.65   &   (1200.10)\\
4  & 962 & 1012   &    9515.58  &   (2141.92) & 8837.39   &   (2150.58)\\
5   & 1166 & 807   &   19354.00   &   (7934.70)& 14138.37   &   (8310.59)\\
        \hline 
    \end{tabular}
    \label{table_ATE}
\end{table}

\begin{table}[h]
    \centering
    \caption{Average treatment effect of 401(k) eligibility on net financial assets. Localized average
    treatment effects are reported by income quintile groups. The regression is estimated by random forest or neural network. Standard errors are reported in parentheses.}
    \begin{tabular}{@{}ccccccc@{}}
    \hline 
        Income quintile & N treated & N untreated & \multicolumn{2}{c}{Random forest}&\multicolumn{2}{c}{Neural network} \\
        \hline 
All & 3682 & 6187   &    8638.15   &   (1621.78)& 7364.66   &   (1844.39)\\
1 & 272 & 1702   &    4874.49   &   (937.86)&4664.61   &   (1309.59)\\
2 & 527 & 1447   &    1957.72   &   (1738.61)&  1635.69   &   (1603.19)\\
3 & 755 & 1219   &    3973.11   &   (1474.72)&5106.03   &   (1287.70)\\
4  & 962 & 1012   &    10056.79  &   (2375.44) & 9529.03   &   (2205.61)\\
5   & 1166 & 807   &   21168.13   &   (8015.79)& 20138.57   &   (7506.92)\\
        \hline 

    \end{tabular}
    \label{table_ATE2}
\end{table}

\subsection{Global and local price elasticity of gasoline demand}

Second, we use our method to estimate the average price elasticity of household gasoline demand: the percentage change in demand due to a unit percentage change in price. This parameter is critical for assessing the welfare consequences of tax changes, and it has been studied in \cite{hausman1995nonparametric,schmalensee1999household,yatchew2001household,blundell2012measuring}. Formally, the parameter of interest is the average derivative of log demand with respect to log price holding income and demographic characteristics fixed. The exact version of this empirical question corresponds to Example~\ref{ex:deriv}. The approximate version of this empirical question corresponds to Example~\ref{ex:policy2}.

We use data from the 1994-1996 Canadian National Private Vehicle Use Survey (\cite{semenova2020debiased_data}), using sample selection and variable construction as in \cite{yatchew2001household} and \cite{belloni2019conditional}. The outcome $Y$ is log gasoline consumption. The variable $D$ with respect to which we differentiate is log price per liter. The raw covariates $X$ are log age, log income, and log distance as well as geographical, time, and household composition dummies. In total we have $n=5001$ observations. We consider the specification $b(D,X)$ previously considered by \cite{semenova2020debiased} augmented with additional interactions. The \cite{semenova2020debiased} specification includes polynomials of continuous covariates, and interactions of log price (and its square) with time and household composition dummies. We further include interactions of log price (and its square) with log age, log age squared, log income, and long income squared to allow for heterogeneity. Altogether, $p=99$.

Table~\ref{table_APE} summarizes results for the entire population and for each quintile of the income distribution. We use $K=5$ folds in cross-fitting. To estimate the RR, we use the generalized Dantzig selector (GDS) procedure introduced in the present work.  To estimate the regression, we use GDS, Lasso, random forest, or neural network. Again, GDS is implemented using the tuning procedure described in Section~\ref{sec:diff}, Lasso is implemented using the tuning procedure described in \citet*{chernozhukov2018learning}, and random forest and neural network are implemented with the same settings as \cite{chernozhukov2018double}. We find average price elasticity of $-0.28$ $(0.06)$ using GDS for both the RR and the regression. Lasso gives similar results. Note that random forest is not differentiable, and the derivative of a neural network may be difficult to extract from a black-box implementation. When using these estimators, we implement a partial difference approximation of the derivative, detailed in Section~\ref{sec:diff}. We conjecture that this approximation explains why the results using random forest appear attenuated and why the results using neural network appear positive or statistically insignificant. Using GDS, we find that localized average price elasticity is statistically significant and negative in each income quintile, with substantial heterogeneity.

\begin{table}[h]
    \centering
     \caption{Estimated average derivative (price elasticity) of gasoline demand. Localized average derivatives are
    reported by income quintile groups. The regression is estimated by GDS, Lasso, random forest, or neural network. Standard errors are reported in parentheses.}
    \begin{tabular}{@{}ccccccccccc@{}}
    \hline 
       Income quintile & N& \multicolumn{2}{c}{GDS} &\multicolumn{2}{c}{Lasso}&\multicolumn{2}{c}{Random forest} & \multicolumn{2}{c}{Neural network}\\
       \hline
        All   &  5001 & -0.28  &   (0.06)&-0.16  &   (0.05) & -0.01   &   (0.06)  &0.15   &   (0.05)\\
        1   & 1001& -0.84   &   (0.13)&-0.44   &   (0.12)&-0.37   &   (0.14)  & 0.06   &   (0.12)\\
        2   & 1000& -0.36   &   (0.12)&-0.27   &   (0.11)&-0.13   &   (0.13)  & 0.42   &   (0.12)\\
        3  & 1000 & -1.40   &   (0.15)&-0.91   &   (0.13)&-0.60   &   (0.13)  &-0.28   &   (0.13)\\
        4   & 1000  & -1.06   &   (0.14)&-0.79   &   (0.14) &-0.32   &   (0.15)  &0.13   &   (0.14)\\
         5  & 1000& -0.11   &   (0.14)&-0.03   &   (0.11)&0.16   &   (0.12)  &0.58   &   (0.10)\\
        \hline 
        
    \end{tabular}
       \label{table_APE}
\end{table}

For comparison, OLS regression of log consumption on log price, log age, log income, and log distance as well as geographical, time, and household composition dummies yields an estimate of $0.14$ $(0.06)$. The linear specification leads to a positive elasticity estimate, contradicting economic intuition (since it says there would be more gasoline consumption when prices are higher). Our localized average price elasticity results using GDS are broadly consistent with \cite{semenova2020debiased}, who more explicitly consider the relationship between average price elasticity and income. In Section~\ref{sec:diff} we report analogous estimates without debiasing. Without debiasing, the GDS and Lasso estimates of quintile elasticities are attenuated due to regularization. The bias is smaller for the random forest and neural network estimates of quintile elasticities.

        

\section{Estimation and inference for high dimensional approximately linear models}\label{sec:estimation}


\subsection{ Best linear approximations for the regression function and the  Riesz representer}
To approximate the regression function, we consider the $p$-vector of dictionary functions  $$x \mapsto b(x) = (b_j(x))_{j=1}^p, \quad b_j \in L^2(F).$$
The dimension $p$ of the dictionary can be large, potentially much larger than $n$.  Let $\Gamma_b$ be the linear subspace
of $L^2(F)$ generated by $b$.   We assume that as $n \to \infty$ we have that $p \to \infty$ and
$
\Gamma_b \to \bar \Gamma:= \text{closure}(\Gamma),
$
where $\bar  \Gamma$ is a linear subspace of $L^2(F)$ with the basis functions $\{\tilde b_j\}_{j=1}^{\infty}$. Here convergence means that any convergent sequence in $\Gamma_b$ has its limit
in $\bar \Gamma$ and for each $\gamma \in \bar \Gamma$ we have a sequence in $\Gamma_b$ converging to it, with respect to the $L^2(F)$ norm. Note that this setup allows the dictionary $b= b_n$ to change with $n$, as for example with $b$-splines.

Here we define $\gamma^\star_0$ as a projection of $Y$ onto $\bar \Gamma$, i.e. $\gamma^\star_0$ is the projection of $Y$ on the infinite set of variables $\{\tilde b_j(X)\}_{j=1}^{\infty}$.  This setup is slightly more general than in the introduction, where
$\gamma^\star_0$ was the conditional expectation function. Of course, if the latter is an element of $\bar \Gamma$, it automatically coincides with $\gamma^\star_0$.

We approximate $\gamma_0^\star$ by the finite-dimensional best linear predictor (BLP) $\gamma_0$ via
 $$
\gamma^\star_0 = \gamma_0 +  r_\gamma := b'\beta_0 + r_\gamma:    \Ep [ b(X) r_\gamma(X) ] =0,
$$
where $r_\gamma$ is the approximation error, and $ \gamma_0 := b'\beta_0$ is the best  linear predictor of $Y$ and best linear approximation to $\gamma^\star_0$. We define $\beta_0$ as a minimal $\ell_1$-norm solution to the system of equations
$$
\min \|\beta\|_1 :  \Ep [ b(X) ( \gamma^\star_0(X) - b(X)'\beta)] = 0,
$$
when $G = \Ep b(X)b(X)'$ is not full rank.

Similarly, we approximate the Riesz representer $\alpha^\star_0$, which exists by Lemma~\ref{lemma: ERR} whenever $L<\infty$, via the best linear approximation $\alpha_0$:
$$
\alpha^\star_0 = \alpha_0 + r_\alpha = b'\rho_0 + r_\alpha: \Ep [r_\alpha (X) b(X) ] = 0.
$$
We define $\rho_0$ as a minimal $\ell_1$-norm solution to the system of equations
$$
\min \| \rho \|_1 : \Ep [(\alpha^\star_0(X) -  b(X)'\rho)b(X)] = 0. 
$$
Using that  $\Ep \alpha_0^\star(X) b(X) = \Ep m(W, b)$, we note that 
\begin{equation}\label{eq:RRP}
0=\Ep [r_\alpha (X) b(X) ] = \Ep  ((\alpha^\star_0(X) -  b(X)'\rho_0)b(X)) =
\Ep  m(W, b)  -  \Ep  \alpha_0(X) b(X).
\end{equation}
Hence $\alpha_0$ is the Riesz representer for $\Ep m(W,\gamma)$ for each $\gamma \in \Gamma_b$.
Here we can interpret $\Gamma_b$ as the collection of test functions on which the representation property (\ref{eq:RRP}) holds.


\begin{definition}[Penultimate and ultimate target parameters] Our  penultimate target is 
the linear functional applied to the BLP $\gamma_0$:
$$
\theta_0 := \Ep [m(W, \gamma_0)] = \Ep  [\alpha_0(X) \gamma_0(X)] = \Ep  [m(W, \gamma_0) +  \alpha_0(X) ( Y - \gamma_0(X)) ].$$
Our ultimate target is the linear functional applied to $\gamma_0^\star$
$$
\theta^\star_0 := \Ep  [m(W, \gamma^\star_0)] =\Ep [\alpha^\star_0(X) \gamma^\star_0(X)]
 = \Ep  [m(W,  \gamma^\star_0) + \alpha^\star_0(X) ( Y -  \gamma^\star_0(X))].$$
\end{definition}
If the approximation errors are such that 
\begin{equation} \label{eq:specified}
	 (\sqrt{n}/\sigma )\int r_\alpha r_\gamma dF \to 0
\end{equation}
our inference will target the ultimate parameter. In the non-regular setup, the second order error condition $\int r_{\alpha}r_{\gamma}dF\leq \sigma/\sqrt{n}$ is weaker than what is usually required for pathwise differentiable functionals (since $\sigma\rightarrow\infty$ is the non-regular case); there is a lower bar for oracle rates in non-regular problems. This phenomenon was also noted by \cite{foster2019orthogonal} and \cite{kennedy2020optimal}. Otherwise our inference will target
an interpretable penultimate parameter.  We  shall formally refer to the latter case as the \textit{misspecified case}.  

\begin{lemma}[Basic properties of the score]\label{lemma:score} Our DML estimator of  $\theta_0$ will be based on the following score function:
$$
\psi (W, \theta; \beta, \rho) = \theta - m(W, b)'\beta - \rho' b(X) (Y - b(X)'\beta),
$$
which has the following properties:
$$
 \partial_{\beta} \psi (W,  \theta; \beta, \rho) = - m(W, b) + {\rho}' b(X) b(X)', \quad \partial_{\rho}\psi(W,  \theta; \beta, \rho) =  -b(X) (Y - b(X)'\beta), 
$$$$
\partial^2_{\beta \beta'} \psi (W,  \theta; \beta, \rho)  =  \partial^2_{\rho \rho'}\psi(W,  \theta; \beta, \rho) =0,
\quad \partial^2_{\beta \rho'}  \psi (W,  \theta; \beta, \rho)   = b(X) b(X)'.
$$
This score function is Neyman orthogonal at $(\beta_0, \rho_0)$:
$$
\Ep [\partial_{\beta} \psi (W,  \theta ; \beta, \rho_0)] =   -\Ep [ m(W, b)] +  G\rho_0 = 0,$$$$
\Ep  [\partial_{\rho} \psi (W,  \theta; \beta_0, \rho)]=  \Ep[  -b(X) (Y - b(X)'\beta_0)] = - \Ep [ b(X) \gamma_0(X)  ]+ G\beta_0 = 0.
$$
\end{lemma}
The second claim of the lemma is immediate from the definition of $(\beta_0, \rho_0)$ and the first follows from elementary calculations. The orthogonality property above says that the score function is  invariant to small perturbations of the nuisance parameters $\rho$ and $\beta$ around their ``true values''
$\rho_0$ and $\beta_0$. This invariance property plays a crucial role in removing the impact of biased estimation of nuisance parameters $\rho_0$ and $\beta_0$ on  the estimation of the main parameters $\theta_0$.

\subsection{Estimators}
Estimation will be carried out using the following Dantzig selector-type estimators (\cite{candes2007dantzig}). In a follow-up work, \citet*{chernozhukov2018learning} consider Lasso-type estimators.

\begin{definition}[Generalized Dantzig selector estimator]   Consider a parameter $t \in T \subset \Bbb{R}^p$, where $T$ is a convex set. Consider the moment functions $t \mapsto g(t) $ and the estimated moment functions $t \mapsto \hat g(t) $,
mapping $\Bbb{R}^p$ to $\Bbb{R}^p$: $$
g(t) = G t - M;   \quad \hat g(t) = \hat G t - \hat M,
$$
where $G$ and $\hat G$ are $p$ by $p$ non-negative-definite matrices and $M$ and $\hat M$
are $p$-vectors.     Define  $t_0$ as a minimal $\ell_1$-norm solution to $g(t) =0$ and assume $t_0 \in T$. Define the GDS estimator $\hat t$ by solving
$$
\hat t \in \arg \min \| t\|_1:  \| \hat g(t) \|_\infty \leq  \lambda, \quad t \in T$$ 
where $\lambda $ is chosen such that 
$ \| \hat g(t_0) -  g(t_0)\|_\infty \leq \lambda,$
with probability at least $1-\epsilon$.
\end{definition}

Here we record the possibility of convex restrictions on the parameter space  by placing $t$ in a convex parameter space $T$. If parameter restrictions are correct, then this can potentially improve theoretical guarantees by weakening the requirements on $G$
and other primitives.

\begin{definition}[GDS for BLP: Dantzig selector]\label{D.1} Given a diagonal positive-definite
normalization matrix $D_{\beta}$, define $\hat \beta_A = D_\beta \hat t$,  where $\hat t$  is the GDS estimator for $t_0=D_{\beta}^{-1}\beta_0$ with
$$
 \quad G = \Ep b(X) b(X)',  \hat  G= \mathbb{E}_A b(X) b(X)',   M =  D_{\beta}^{-1}\Ep Y b(X),   \hat M  = D_{\beta}^{-1}\mathbb{E}_A  Y b(X);  T_\beta \subset \Bbb{R}^p.
$$
\end{definition}
In this setting, our estimator specializes to the original Dantzig selector. In practice, we use $T_\beta = \Bbb{R}^p$, although when we are interested in average derivative functionals, it is theoretically
helpful to impose the convex restrictions of the sort $T = \{ t \in \Bbb{R}^p: \sup_{x \in \mathcal{X}} | \partial_d b(x)'t | \leq B\}$, where $B$ is some  a priori known upper bound on the derivative.  Ideally, $D_{\beta}$ is chosen such that $\mathrm{diag}( Var( D^{-1}_{\beta} (\hat G \beta_0 - \hat M) ) = I$.  Our practical algorithm given in Section~\ref{sec:diff} estimates $D_\beta$ from the data.

\begin{definition}[GDS for Riesz representer]\label{D.2} Given a diagonal positive-definite
normalization matrix $D_{\rho}$, define $\hat \rho_A = D_{\rho}\hat t$,  where $\hat t$  is the GDS estimator 
of the parameter $t_0=D_{\rho}^{-1}\rho_0$ with
$$
\quad G = \Ep b(X) b(X)',   \hat G = \mathbb{E}_A  b(X) b(X)',  M = D_{\rho}^{-1}\Ep m(W, b), \hat M  =  D_{\rho}^{-1}\mathbb{E}_A  m(W, b); T_\rho \subset \Bbb{R}^p.
$$
\end{definition}
In this setting, our estimator is a generalization of the original Dantzig selector. In practice, we are using $T_\rho =\Bbb{R}^p$, even though it is possible to exploit some structured restrictions
on the problem motivated by the nature of the universal Riesz representers. Ideally, $D_{\rho}$ is chosen such that $\mathrm{diag}( Var( D^{-1}_{\rho} (\hat G \rho_0 - \hat M) ) = I$. Our practical algorithm given in Section~\ref{sec:diff} estimates $D_\rho$ from the data.

We now define the DML estimator with Riesz Representers, which makes use of cross-fitting.
\begin{definition}[DML with RR] Consider the partition of $\{1,...,n\}$ into $K \geq 2$ blocks $(I_k)_{k=1}^K$, with $m= \lfloor n/K\rfloor$
observations in $I_k$, for $k< K$ and $\lceil n/K \rceil $ remaining in $I_K$. For each $k =1,...,K$, let
$ \hat \beta_{k}$ and $\hat \rho_{k}$ denote GDS estimators obtained using data $(W_i)_{i \in I^c_k}$, where $I^c_k= \{1,...,n\} \setminus I_k$, and let estimator $\hat \theta_{k}$ be defined as
$$
\hat{\theta}_k= \mathbb{E}_{I_k} [m(W, b)'\hat \beta_k + \hat \rho_{k}' b(X) (Y - b(X)'\hat \beta_k)].$$
Define the DML estimator $\hat \theta$ as
the average:
$$
\hat \theta = \sum_{k=1}^{K} \hat \theta_{k} w_k;  \quad  w_k = \frac{\lfloor n/K \rfloor}{n} \text{ if } k < K, \quad w_K = 
\frac{\lceil n/K\rceil }n .
$$
\end{definition}

\subsection{Properties of DML: Main result}\label{sec:DML}
We provide a single non-asymptotic result that allows us to cover both global and local functionals, implying uniformly
valid rates of concentration and normal approximations over large sets of $P$.

Consider the oracle estimator based upon the true score functions:
$$
\bar \theta := \theta_0 - n^{-1} \sum_{i =1}^n \psi_0(W_i), \quad \psi_0(W) :=\psi(W, \theta_0; \beta_0, \rho_0).
$$
We seek to establish minimal conditions under which the DML estimator approximates the oracle estimator, and is approximately
normal with distribution 
$$
N(0, \sigma^2/n), \quad \sigma := \| \psi_0\|_{P,2}.
$$
For regular functionals $\sigma$ is bounded, giving $1/\sqrt{n}$ concentration around $\theta_0$, and for non-regular functionals $\sigma \propto L \to \infty$ requring $L/\sqrt{n} \to 0$ to get concentration.  Our normal approximation is accurate if kurtosis of $\psi_0$ does not grow too fast: 
$$
(\kappa/\sigma)^3 /\sqrt{n} \text{ is small},  \ \ \kappa:= \| \psi_0\|_{P,3}.
$$
In the regular case $(\kappa/\sigma)^3$ is bounded, but for the non-regular cases it can scale as fast as $L$, again requiring $L/\sqrt{n} \to 0$.

 Fix all of these sequences and the constants.  
Define the guarantee set:
$$
\mathsf{S} = \left\{ \begin{array}{c}(u, v) \in \Bbb{R}^{2p}:  \sqrt{ u'Gu} \leq r_1, \sqrt{ v'Gv} \leq \sigma r_2,  \ \
|u'Gv| \leq \sigma r_3, \beta_0 + u \in T_\beta,  \rho_0 + v \in T_{\rho} \end{array} \right \},
$$
We will take $u=\hat{\beta}_k-\beta_0$ and $v=\hat{\rho}_k-\rho_0$. As such, $r_1$ measures the non-asymptotic mean square rate for the BLP; $r_2$ measures the non-asymptotic mean square rate for the RR; and $r_3$ measures how the estimation errors interact. Note the presence of $\sigma$ acting on $r_2$ and $r_3$, which accommodates non-regular functionals. We will instantiate $(r_1,r_2,r_3)$ as fast and slow rates by analyzing the GDS estimator, in Theorem~\ref{theorem: rates1} below.

 Next, define
$\mu$ to be the smallest modulus of continuity such that on $(u,v) \in \mathsf{S} $
$$
\sqrt{Var} ( (-m(W, b) + {\rho_0}' b(X) b(X)) 'u) \leq  \mu \sigma  \| b'u\|_{P,2},
\sqrt{Var} ( (Y - b(X)'\beta_0) b(X) 'v) \leq \mu \| b'v\|_{P,2}, \\
$$$$
 \sqrt{Var} (u 'b(X) b(X)' v) \leq \mu ( \| b'u\|_{P,2}+\| b'v\|_{P,2}).
$$
In typical applications, the modulus of continuity $\mu$ is bounded.  
Indeed, if elements of the dictionary are bounded with probability one, $ \| b(X) \|_\infty \leq C$, then we can select $\mu = C B$ for many functionals of interest, so the assumption is plausible. If $b(X) = X$ are sub-Gaussian, then this assumption is also easily satisfied; however, this case is not of central interest to us. See \cite{chernozhukov2021simple} for a more general discussion.

Consider $P$ that satisfies the following conditions. 
\begin{itemize}

\item[R$(\delta)$]\textit{With probability $1-\varepsilon$, the estimation errors $\{(\hat \beta_k- \beta_0, \hat \rho_k- \rho_0)\}_{k=1}^{K}$ take values in $\mathsf{S}^K$},
\textit{with quality of the guarantee obeying} $$ \sigma^{-1}  ( \sqrt{m}\sigma r_3  +  \mu r_1 (1+ \sigma)  + \mu \sigma r_2 )  \leq \delta.$$

\end{itemize}
$R(\delta)$ is a requirement on how the sequences $(r_1,r_2,r_3)$ evolve relative to $(\sigma,\mu,m)$. We will formally verify $R(\delta)$ for the approximately sparse setting, in Corollary~\ref{Cor:SUFF} below. $R(\delta)$ is the key condition for our main result, Theorem~\ref{theorem: DML}.

\begin{theorem}[Adaptive estimation and approximate Gaussian inference]\label{theorem: DML} Suppose $K$ divides $n$ for simplicity. Under condition R$(\delta)$,  we have the adaptivity property, namely the difference
between the DML and the oracle estimator is small: for any $\Delta \in (0,1)$,
$$
 |\sqrt{n}(\hat \theta - \bar \theta)/\sigma| \leq  \sqrt{K} 4 \delta/\Delta
$$
with probability at least $1- \varepsilon- \Delta^2$.

As a consequence, $\hat \theta$ concentrates in a $\sigma/\sqrt{n}$ neighborhood of $\theta_0$, with deviations approximately distributed according to the Gaussian law $\Phi(z)=  \Pr(N(0,1) \leq z)$:
\begin{eqnarray*}
& & \sup_{z \in \Bbb{R}}\Big | \Pr ( \sigma^{-1} \sqrt{n} (\hat \theta_0 - \theta_0) \leq z) - \Phi(z) \Big |  \leq A(\kappa/\sigma)^3 n^{-1/2}+  \sqrt{K} 2 \delta/ \Delta  + \varepsilon+ \Delta^2,
 \end{eqnarray*}
where $A < 1/2$ is the sharpest absolute constant in the Berry--Esseen bound.
\end{theorem}

The conclusions of this result are distinguished from those of \cite{chernozhukov2018double} and \citet*{chernozhukov2018learning} in applying to local, nonparametric objects, in providing finite sample bounds, and in being uniform over the parameter space. The conclusions are similar to this previous work in relying on a rate condition that is the product of rates of estimation for two distinct functions, here the regression and the Riesz representer.

The constants can be chosen  to  yield an asymptotic result.

\begin{corollary}[Uniform asymptotic adaptivity and Gaussianity]\label{corollary:uniform} Let $\mathcal{P}_n$ be any nondecreasing set of probability laws $P$ that obey condition $R(\delta_n)$  where $\delta_n \to 0$ is a given sequence. Then the DML estimator $\hat \theta$ is uniformly asymptotically equivalent to the oracle estimator $\bar \theta$,
that is $$|\sqrt{n}(\hat \theta - \bar \theta)/\sigma| = O_P (\delta_n)$$ uniformly in $P \in \mathcal{P}_n$ as $n \to \infty$.
In addition, if for each $P \in \mathcal{P}_n$ the kurtosis of $\psi_0$ does not grow too fast, namely:
$$
(\kappa/\sigma)^3/\sqrt{n} \leq \delta_n,
$$
we have that $\sqrt{n} (\hat \theta - \theta_0)/\sigma$ is asymptotically Gaussian uniformly in $P \in \mathcal{P}_n$:
$$
\lim_{n \to \infty} \sup_{P \in \mathcal{P}_n} \sup_{z \in \Bbb{R}}\Big | \Pr_P (\sqrt{n}(\hat \theta_0 - \theta_0)/\sigma \leq z) -  \Phi(z) \Big | = 0.
$$
\end{corollary}

Hence the DML estimator of the linear functionals of the BLP function $\gamma_0$ enjoys good properties under the stated regularity conditions. This result does not distinguish between inference on global functionals from inference on local functionals, as long as the latter are not perfectly localized. We state a separate result for perfectly localized functionals below.


\begin{corollary}[Inference on the ultimate parameter $\theta_0^\star$] Suppose 
that, in addition to conditions of Corollary~\ref{corollary:uniform},  $P$ satisfies the small approximation error condition:
\begin{equation}\label{eq: smallbias}
(\sqrt{n}/\sigma) | \theta_0 - \theta^\star_0| = (\sqrt{n}/\sigma)   \left |\int r_\alpha r_\gamma dF \right | \leq \delta.
\end{equation}
Then conclusions of Theorem~\ref{theorem: DML} hold with  $\theta_0^\star$ replacing $\theta_0$, with  
$\sqrt{K} 4 \delta/\Delta$ increased by $\delta$, and the same probability. Conclusions of Corollary~\ref{corollary:uniform} continue to hold
with $\theta_0^\star$ replacing $\theta_0$ for a class of probability laws $\mathcal{P}_n$, provided each $P \in \mathcal{P}_n$ satisfies the conditions of Corollary~\ref{corollary:uniform} and (\ref{eq: smallbias}) for the given $\delta= \delta_n \to 0$.
\end{corollary}

The approximation bias for the ultimate target can be plausibly small due to the fact that many rich function classes admit regularized linear approximations with respect to conventional dictionaries $b$.  For instance, \cite{tsybakov2012introduction} and \cite{belloni2014pivotal} show small approximation bias using Fourier bases as dictionaries, and using Sobolev and rearranged Sobolev balls, respectively, as the function classes. 

\begin{corollary}[Inference on the perfectly localized parameter] Suppose 
that, in addition to conditions of Corollary~\ref{corollary:uniform},  $P$ satisfies the small approximation error condition:
\begin{equation}\label{eq: PLC1}
\sqrt{n}| \theta_0 (\gamma_0;  \ell_h) - \theta_0(\gamma_0^\star; \ell_h) |/\sigma  = \sqrt{n}\left |\int r_\alpha r_\gamma dF \right |/\sigma \leq \delta,
\end{equation}
and  the localization bias is small:
\begin{equation}\label{eq: PLC2}
\sqrt{n}|\theta_0(\gamma_0^\star; \ell_h)  -\theta_0(\gamma_0^\star; \ell_0) |/\sigma \leq \delta,
\end{equation}
Then conclusions of Theorem~\ref{theorem: DML} hold with  $\theta_0(\gamma_0^\star; \ell_0)$ replacing $\theta_0$, with  
$\sqrt{K} 4 \delta/\Delta$ increased by $2\delta$, and the same probability. Conclusions of Corollary~\ref{corollary:uniform} continue to hold
with $\theta^\star_0(\gamma^\star; \ell_0)$ replacing $\theta_0=\theta_0 (\gamma_0;  \ell_h)$ for a class of probability laws $\mathcal{P}_n$, provided each $P \in \mathcal{P}_n$ satisfies the conditions of Corollary~\ref{corollary:uniform} 
and (\ref{eq: PLC1})-(\ref{eq: PLC2}) for the given $\delta= \delta_n \to 0$.
\end{corollary}

\subsection{Semi-parametric efficiency}

Below we use concepts from semi-parametric efficiency, as presented in
\cite{bickel1993efficient}  and 
\cite{vaart}; we do not recall them here for brevity.

The DML estimator $\hat \theta$ will be asymptotically efficient for estimating 
 $\theta^\star_0$, defined as a functional of $\gamma^\star_{0}$, the projection
 of $Y$ on $\bar{\Gamma}$. The distribution of a data observation is unrestricted in
this case, so that there will only be one influence function for each
functional of interest, and the estimator is asymptotically linear with that
influence function.  The standard semiparametric efficiency results then imply
that our estimator will have the smallest asymptotic concentration among
estimators that are locally regular; see \cite{bickel1993efficient}  and 
\cite{vaart}.  

Our formal result stated below only implies efficiency for the regular case, where 
the operator norm of the function $L$ is bounded, holding $P$ fixed.  We expect
that a similar result continues to hold with $L \to \infty$, by developing an appropriate
formalization that handles $P$ changing with $n$ and rules out super-efficiency phenomena. However,
this formalization requires a separate major development, which we leave to future research.
In what follows, the notation $\gamma^\star_{0,P}$ emphasizes the dependence of the projection $\gamma^\star_0$ on $P$.


\begin{theorem}[Efficiency]\label{theorem:eff} Let $
\psi_{0}^\star(W) := \theta_0^\star- m(W, \gamma_0^\star)-\alpha^\star _0(X)(Y- \gamma_0^\star(X)).
$
Suppose that $\Ep[Y^{2}]<\infty,$ $\Ep[\psi_0(W)^{2}]<\infty,$ and $m(W,\gamma)$ is mean square continuous in $\gamma$ under $P$.  Then $\theta_{0,P} := \int m(w, \gamma^\star_{0,P}) dP(w)$
is differentiable at $P$, in the sense that
$$
\lim_{\tau \searrow 0 } \frac{\theta_{0,P_\tau} - \theta_{0,P}}{\tau}  = \Ep_{P} \delta(W) \psi_0(W),
$$
where  $\psi_0$ is called the influence function and is unique, and the directional perturbation $P_{\tau}$ is defined
as  $dP_{\tau} = dP[1+ \tau \delta]$, where the direction $\delta$ is any element of the tangent set $\{\delta \text{ measurable } : \mathcal{W} \to \Bbb{R}: \int \delta dP =0, \|\delta\|_\infty< M\}$ for each $0<M<\infty$.   Consequently, the asymptotic variance of every regular sequence of estimators is bounded below by $\| \psi_0\|_{P,2}.$ 
Further, since the tangent set is a convex cone, other conclusions of the Theorem~25.20 and Theorem~25.21 of \cite{vaart} also hold, namely the convolution and the minimax characterization of the efficiency.  

  \end{theorem}

\subsection{Properties of GDS estimators} 


Our goal is to verify that the guarantee $R(\delta)$ holds. In particular we have to analyze $(r_1,r_2,r_3)$ by bounding the population prediction norm 
$
v \mapsto \sqrt{v' G v}.
$
This is a more nuanced problem than bounding the empirical prediction norm $v \mapsto \sqrt{v' \hat G v}$, which has been
accomplished in a variety of prior analyses done on Dantzig-type and Lasso-type estimators.

We begin with the following condition, which only controls the max of error rates  and controls the $\ell_1$ norm of true parameters:
 \begin{itemize}
\item[MD] \textit{We have that $t_0 \in T$ and $\|t_0\|_1 \leq B$, where $B \geq 1$, and the empirical moments obey the following bounds
 with probability  at least  $1 - \varepsilon$, for $\bar \lambda \geq \lambda$ $$
 \| \hat G - G\|_\infty \leq \bar \lambda, \ \| \hat G t_0 - \hat M\|_{\infty}\leq \lambda.
$$}
\end{itemize}

The bounds on $\ell_1$ norm of coefficients are naturally motivated, for example, by working in Sobolev or rearranged Sobolev
spaces (see, \cite{tsybakov2012introduction} and \cite{belloni2014pivotal}, respectively). Rearranged Sobolev spaces allow the first $p$ regression coefficients in the series expansion to be arbitrarily rearranged, allowing a much greater degree of oscillatory behaviors than in the original Sobolev spaces. The complexity of these function classes are also different. 
Sobolev spaces are Donsker sets under sufficient smoothness, whereas rearranged Sobolev spaces have the covering
entropy bounded below by $\log p$ and are not Donsker if $p \to \infty$.

At the core of this approach is the restricted set
$$
 S (t_0, \nu)  := \{ \delta: \| G \delta \|_\infty \leq \nu , \| t_0+ \delta\|_1 \leq \|t_0\|_1, t + \delta \in T\},
$$
where $\nu$ is the noise level. As demonstrated in the proof, the GDS estimator belongs to this set with high probability $1- \epsilon$ for 
the noise level $\nu= 4B \bar \lambda,$ 
where $\lambda$ is the penalty level of GDS ($\nu$ scales like $\sqrt{ \log (p \vee n) }/\sqrt{n}$ in our problems).

\begin{definition}[Effective dimension]
Define the effective dimension of $t_0$ at the noise level $\nu>0$ as:
$$
s(t_0) := s(t_0; \nu) := \sup_{\delta \in  S (t_0, \nu) } | \delta'G \delta|/\nu^2.
$$
\end{definition}
The effective dimension is defined in terms of the population (rather than sample) covariance matrix $G$, which makes it easy to verify regularity conditions.  Note that if  $G=I$ and $\|t_0\|_0 =s$, then $s(t_0) \leq s.$
More generally, $s(t_0)$ measures the effective difficulty of estimating $t_0$ in the prediction norm, created by design $G$ and the structure of $t_0$.
The condition imposes no conditions on the restricted or sparse eigenvalues of $G$. For example,
take $G = 11'$, a rank 1 matrix, and suppose $\|t_0\|_0 = 1$.  Then $s(t_0) \leq 1$ holds in this case, giving 
useful and intuitive performance bounds, while the standard restricted eigenvalues and cone invertibility factors are all zero
in this case, yielding no bounds on the performance in the population prediction norm.   This type of example illustrates
the possibility of accommodation of  overcomplete (multiple or amalgamated) dictionaries in $b$, whose use in conjunction with $\ell_1-$ penalization has been advocated by \cite{donoho2005stable}.  Of course, the bounds
on effective dimension follow from the bounds on cone-invertibility factors and restricted eigenvalues.

Given a vector $\delta \in \Bbb{R}^p$,  let $\delta_A$ denote a vector with 
the $j$-th component set to $\delta_j$ if $j \in A$ and $0$ if $j \not \in A$. 

\begin{lemma}[Bound on effective dimension in approximately sparse model]\label{lemma: boundS}  Suppose that 
$t_0$ is approximately sparse, namely $$|t_{0}|_j^* \leq A j^{-a} \quad j =1,...,p,$$
for some finite positive constants $A$ and $a>1$, where $(|t_{0}|_j^*)_{j=1}^{p}$ is the non-increasing rearrangement of
$(|t_{0j}|)_{j=1}^p$. Let $ t_0^{\mathcal{M}} := t_0 (1 ( |t_0| > \nu) := (t_{0j} 1(| t_{0j}| >\nu ))_{j=1}^p$ denote the vector with components smaller than $\nu$
trimmed to 0. Then
$$
s(t_0; \nu) \leq  s\times \left(k^{-1} \vee \frac{6a}{a-1}\right ), \quad   \| t_0^{\mathcal{M}}\|_0 \leq s:= (A/\nu)^{1/a},
$$
$k$ is the cone invertibility factor: $$
k := \inf \frac{  |\mathcal{M}|  \|G\delta\|_\infty}{\|\delta\|_1}:  \delta \neq 0, \ \ \|\delta_{\mathcal{M}^c}\|_1 \leq 2 \|\delta_{\mathcal{M}}\|_1,$$ $\mathcal{M} = \text{support} ( t_0^{\mathcal{M}})$, $\mathcal{M}^c = \{1,...,p\} \setminus \mathcal{M}$, and $|\mathcal{M}|\leq s$. \end{lemma}
 The cone invertibility factor is a generalization of the restricted eigenvalue condition of \cite{bickel2009simultaneous}, proposed by \cite{ye2010rate}. The concept of the effective dimension does not split $t_0$ into a sparse component and a small dense component, as is done in the now standard analysis of $\ell_1$-regularized estimators of approximately sparse $t_0$.   The effective dimension is simply stated in terms of $t_0$ alone.

Since approximate sparsity is a simple condition that implies a bound on effective dimension, we pause and interpret approximate sparsity in the context of a motivating example from causal inference. In particular, we revisit ATE (Example~\ref{ex:ate}). For simplicity, consider the global parameter and assume that the function $\Ep[Y|D,Z]$ is an element of $\bar{\Gamma}$, so that $\gamma_0^{\star}(D,Z)=\Ep[Y|D,Z]$ and $\alpha_0^{\star}(D,Z)=D/\pi_0^{\star}(Z)-(1-D)/(1-\pi_0^{\star}(Z))$ where $\pi_0^{\star}(Z)=\Ep[D|Z]$ is the propensity score. Consider the dictionary $b(d,z)=(dq(z)',(1-d)q(z)')'$ where $\{q_j(z)\}_{j=1}^{p/2}$ are the initial $p/2$ elements of a sequence of basis functions that approximates the functions $\Ep[Y|1,Z]$, $\Ep[Y|0,Z]$, $1/\pi_0^{\star}(Z)$, and $1/(1-\pi_0^{\star}(Z))$. 

Suppose the minimal $\ell_1$-norm mean square projections of $\Ep[Y|1,Z]$ and $\Ep[Y|0,Z]$ onto $\{q_j(z)\}_{j=1}^{p/2}$ are approximately sparse after rescaling appropriately by $D^{-1}_{\beta}$. (Note that if $\Ep[Y|1,Z]$ and $\Ep[Y|0,Z]$ are already approximately sparse then so are their projections.) It follows that the minimal $\ell_1$-norm mean square projection of $\gamma_0^{\star}$ is approximately sparse and $s_{\beta}:=s(D^{-1}_\beta \beta_0;\nu)$ is small. 

Suppose instead that the minimal $\ell_1$-norm mean square projections of $1/\pi_0^{\star}(Z)$ and $1/(1-\pi_0^{\star}(Z))$ onto $\{q_j(z)\}_{j=1}^{p/2}$ are approximately sparse after rescaling appropriately by $D^{-1}_{\rho}$. (Note that if $1/\pi_0^{\star}(Z)$ and $1/(1-\pi_0^{\star}(Z))$ are already approximately sparse then so are their projections.)  It follows that the minimal $\ell_1$-norm mean square projection of $\alpha_0^{\star}$ is approximately sparse and $s_{\rho}:=s(D^{-1}_\rho \rho_0;\nu)$ is small.

\begin{lemma}[Non-asymptotic bound for GDS in population prediction norm]\label{lemma:RMD} Suppose that MD holds.
Then with probability $1- 2\varepsilon$ the estimator $\hat t$ exists and obeys:
$$
(\hat t-t_0)'G(\hat t- t_0) \leq  ( s(t_0; \nu) \nu^2) \wedge (2B \nu).
$$

\end{lemma}

The bound is a minimum of what is called the ``fast rate bound'' and the ``slow rate'' bound. This result tightens the result in \cite{chatterjee2013assumptionless} who established a ``slow rate'' bound (in the context of Lasso) that applies under no assumptions on $G$. If the effective dimension is not too big, as in the examples above, the ``fast rate'' $s(t_0;\nu) \nu^2$ provides a tighter bound under weak assumptions on $G$. It is important to emphasize that the result is stated in terms
of the population prediction norm rather than the empirical norm.

We now apply this result to GDS estimators of the Riesz representer and the BLP.
We impose the following conditions. Let $\Bbb{G}_A$ denote the empirical process over $f \in \mathcal{F}: \mathcal{W} \to \Bbb{R}^p$ and $ i \in A$, namely $$\mathbb{G}_A f := \mathbb{G}_A f (W) := | I | ^{-1/2} \sum_{ i \in A} (f(W_i) - P f), \quad Pf :=  P f(W) := \int f(w) d P(w).$$

The following is a sufficient condition that will deliver the guarantee $R(\delta)$ for $\delta \to 0$.  Let $\tilde \ell$ denote
a positive constant (that increases to $\infty$ as $n \to \infty$ in the asymptotic results).

\begin{itemize}

\item[SC] \textit{(a)  The $\ell_1$ norms of coefficients are bounded as $\|D^{-1}_\rho \rho_0 \|_1 \leq B$ and  $\|D^{-1}_\beta \beta_0\|_1 \leq B$, for $B \geq 1$, and the scaling matrices obey $ \|D_{\rho} v\| \leq \mu_D \sigma \| v\| $ for $ D^{-1}_\rho v \in S ( D^{-1}_\rho \rho_0, \nu)$
and $ \|D_{\beta} u\| \leq \mu_D \| u\| $ for $D^{-1}_\beta u \in S (D^{-1}_\beta \beta_0, \nu)$ for  $\nu = 4 B \tilde \ell/\sqrt{n}$.   (b) Given a random subset $A$ of $\{1,...,n\}$ of size $m \geq n - \lfloor n/K \rfloor $, dictionary $b$ obeys with probability at least $1-\epsilon$,}
$
\|\Bbb{G}_{A} b b'\|_\infty \leq \tilde \ell. 
$ \textit{ (c) The penalty levels $\lambda_{\rho}$ and $\lambda_{\beta}$
are chosen 
  such that with probability at least $1- \epsilon$,
  $
  \| D_{\beta}^{-1} (\Bbb{G}_{A} b b' \beta_0 - \Bbb{G}_{A} Y b(X))\|_\infty/\sqrt{m}   \leq \lambda_\rho,$ $
  \| D_{\rho}^{-1} (\Bbb{G}_{A} b b' \beta_0 - \Bbb{G}_{A} m(W, b))\|_\infty/\sqrt{m}  \leq \lambda_\beta,
    $
and are not overly large, $ \lambda_{\beta} \vee  \lambda_{\rho} \leq  \tilde \ell/\sqrt{m}$.}

\end{itemize}

SC(a)  records a restriction on the $\ell_1$ norm of $\beta_0$ and $\rho_0$. For instance,
in Examples~\ref{ex:ate},~\ref{ex:policy1}, and~\ref{ex:policy2}, $D_\rho \asymp \sigma I  \asymp L I$, which requires the $\ell_1$-norm of $\rho_0$ to increase at most at 
the speed $L\asymp \sigma$. 

SC(b) is a weak assumption: the bound $\bar \lambda $ and the penalty level $ \lambda$ can be chosen proportionally to  $\sqrt{ \log (p \vee n) }/\sqrt{n}$,  that is $$\tilde \ell \asymp \sqrt{\log (p \vee n)}$$ using self-normalized moderate deviation bounds (\cite{jing2003self,belloni2014pivotal}) or high-dimensional central limit theorems (\cite{chernozhukov2017central}), under mild moment conditions, without requiring sub-Gaussianity.   For instance, \cite{belloni2014pivotal} employ these tools to show that, for the bounded design case $\| b\|_\infty \leq C$, $\lambda$ can be chosen as in the Gaussian error case, provided that errors follow $t(2+\delta)$ distribution (having above $2$ bounded moments), and get the error bounds similar to the Gaussian case.  Here we state a general condition as our working assumption, instead of focusing on more specific condition that get us Gaussian-type conclusions.

\begin{theorem}[GDS for BLP and RR]\label{theorem: rates1}  Suppose SC holds. Then with probability at least $1- K 4\epsilon$,
we have that $
u = \hat \beta_A - \beta_0$ and $v =  \hat \rho_A - \rho_0
$ obey, for some absolute constant $C$,
$$
u'G u \leq r_1^2, \ \ v'Gv \leq  \sigma^2 r^2_2 ,  \ \ |u'Gv|  \leq \sigma r_3 ,
$$$$
r_1^2 =  C \mu_D^ 2  (B^2 \tilde \ell^2  s_\beta/ n)  \wedge  ( B^2 \tilde \ell /\sqrt{n} ), \quad r_2^2  =   C \mu_D^ 2   ( B^2 \tilde \ell^2 s_{\rho} / n)  \wedge  ( B^2 \tilde \ell /\sqrt{n}), \quad r_3 = r_1 r_2,
$$
where $s_\beta$ and $ s_\rho$ are the effective dimensions for parameters $D^{-1}_\beta \beta_0$ and $D^{-1}_\rho \rho_0$ for the noise level $\nu= 4 B \tilde \ell/\sqrt{n}$
 \end{theorem}
 
 In other words, we have instantiated $(r_1,r_2,r_3)$ for approximately sparse models in the guarantee set $\mathsf{S}$. We have the following corollary, which verifies $R(\delta)$ for approximately sparse models and hence provides sufficient conditions for Theorem~\ref{theorem: DML}.
 
 \begin{corollary}[Sufficient condition for $R(\delta)$]\label{Cor:SUFF}  Suppose SC holds.  The guarantee $R(\delta)$ holds with $\varepsilon = 1 - K 4 \epsilon$, provided
$$
\text{either } C s_\beta \leq \sqrt{n}\delta^2/(\tilde \ell^3 \mu^2 \mu^2_D) \text{ or } C  s_\rho  \leq  \sqrt{n} \delta^2/(\tilde \ell^3 \mu^2 \mu^2_D),
$$
for some large enough constant $C$ that only depends on $B$ and $K$.
\end{corollary}

\begin{remark}[Sharpness of conditions: Double sparsity robustness]  This gives sufficient conditions such that (ignoring
slowly growing term $\tilde \ell$) the condition $R(o(1))$ holds if
$$
\text{ either $s_\beta \ll \sqrt{n} \ \text{ or }  s_\rho \ll \sqrt{n}$,}
$$ 
where $s_\beta$ and $s_\rho$ are measures of the effective dimensions of parameters $D_\beta^{-1} \beta_0$ and $D_{\rho}^{-1}\rho_0$. In well-behaved exactly sparse models, these effective dimensions are proportional to the sparsity indices divided by restricted eigenvalues.  The latter possibility allows one of the parameter values to be ``dense'', having unbounded effective dimension, in which case this parameter can be estimated at some ``slow''  rate $n^{-1/4}$. 
These types of conditions appear to be rather  sharp, matching similar conditions  used in \cite{javanmard2018debiasing} in the case of inference on a single coefficient in Gaussian exactly sparse linear regression models.
\end{remark}

\section{Estimation and inference using general regression learners}\label{sec:general}

In this section we generalize the previous analysis to allow for any
regression learner $\hat{\gamma}$ of $\Ep[Y|X]$ to be used in the construction
of the estimator. As we have done in preceding sections we continue to include
local functionals in our analysis, so that the results apply to nonregular
objects as well as regular ones that can be estimated $\sqrt{n}$-consistently. 

Compared to the global case, the local case may have smaller regularization and model selection biases relative to the variance. Nonetheless, bias correction is important for inference in theory and in practice. Theoretically, the local case begins to resemble the global case as the number of dimensions being integrated increases. Empirically, we provide local estimates without bias correction in Section~\ref{sec:diff}. The differences can be substantial.

The only conditions we will impose on the regression learner are certain $L^{2}$
convergence properties that we will specify in this section. These properties
will allow for a wide variety of learners, including GDS, Lasso, neural
nets, boosting, and others. Thus we provide estimators of local functions that
can be constructed using many regression learners.

We continue to consider estimators that use cross-fitting and have the form%
$$
\hat{\theta}=\frac{1}{n}\sum_{k=1}^{K}\sum_{i\in I_{k}}\left\{  m(W_{i}%
,\hat{\gamma}_{k})+\hat{\alpha}_{k}(X_{i})[Y_{i}-\hat{\gamma}_{k}%
(X_{i})]\right\}  .
$$
where $\hat{\gamma}_{k}$ denotes the regression learner computed from
observations not in $I_{k}$ and $\hat{\alpha}_{k}(x)=b(x)^{\prime}\hat{\rho
}_{k}$ is the GDS learner of the Riesz representer described in previous sections.

To allow for as many regression learners as possible under as weak conditions
as possible we focus on asymptotic analysis in this section. The fundamental
property we will require of $\hat{\gamma}$ is that it have some mean square
convergence rate as an estimator of the true conditional mean $\gamma
_{0}^{\star}.$ Specifically we require that there is $r_{1}^{\star}$ converging
to zero such that for each $k,$
\begin{equation}
\left\Vert \hat{\gamma}_{k}-\gamma_{0}^{\star}\right\Vert _{P,2}=O_{p}%
(r_{1}^{\star}).\label{L2rate}%
\end{equation}
For purposes of formulating regularity conditions it is also useful to work
with $\alpha_{0}^{\star}$ rather than $\alpha_{0}$. We will also require that%
\begin{equation}
\left\Vert \alpha_{0}^{\star}(\hat{\gamma}_{k}-\gamma_{0}^{\star})\right\Vert
_{P,2}=o_{p}(\sigma),\text{ }\left\Vert m(\cdot,\hat{\gamma}_{k}-\gamma
_{0}^{\star})\right\Vert _{P,2}=o_{p}(\sigma).\label{L2cons}%
\end{equation}
In the regular case these conditions generally follow from the mean square
consistency of $\hat{\gamma}_{k}$ under boundedness of $\alpha_{0}^{\star}.$ In
nonregular cases they may impose additional conditions. For example, under the
conditions of Lemma~\ref{lemma: lstr13} it will be sufficient for these conditions to hold that%
$$
h^{-p_{1}/2}\left\Vert \hat{\gamma}_{k}-\gamma_{0}^{\star}\right\Vert
_{P,2}\to_{\Pr}0.
$$
This condition will hold as long as $h$ grows slowly enough relative to the
mean-square convergence rate of each $\hat{\gamma}_{k}.$

Recall from Theorem~\ref{theorem: rates1} that $r_{2}$ is the convergence rate of $\sigma
^{-1}\left\Vert \hat{\alpha}_{k}-\alpha_{0}\right\Vert _{P,2}$. Let
$r_{2}^{\star}=r_{2}+\sigma^{-1}\left\Vert \alpha_{0}-\alpha_{0}^{\star
}\right\Vert _{P,2}$ and%
$$
\psi_{0}^{\star}(W)=\theta_{0}-m(W,\gamma_{0}^{\star})-\alpha_{0}^{\star
}(X)[Y-\gamma_{0}^{\star}(X)].
$$

\begin{theorem}[Asymptotic Gaussian inference with general regression learner]\label{theorem: asymptotic} 
Suppose $Var(Y|X)$ is bounded; $r_{1}^{\star}\rightarrow0$ and $r_{2}^{\star}\rightarrow
0$; equations (\ref{L2rate}) and (\ref{L2cons}) are satisfied; and $\sqrt{n}r_{1}^{\star}r_{2}^{\star}\rightarrow0$. Then
$$
\hat{\theta}=\theta_{0}-\frac{1}{n}\sum_{i=1}^{n}\psi_{0}^{\star}(W_{i}%
)+o_{p}\left(\frac{\sigma}{\sqrt{n}}\right)
,\quad \text{hence}\quad
\frac{\sqrt{n}}{\sigma}(\hat{\theta}-\theta_{0})\overset{d}{\rightarrow}\mathcal{N}(0,1).
$$
\end{theorem}

This result shows that asymptotic linearity of the estimator $\hat{\theta}$
will result if $r_{2}^{\star}\rightarrow0$ fast enough relative to
$r_{1}^{\star}$. Asymptotic linearity implies asymptotic Gaussian inference by standard central limit theorem arguments. As in regular doubly robust estimation problems it allows for
a tradeoff between the speed of convergence $r_{2}^{\star}$ of the Riesz
representer and $r_{1}^{\star}$ of the regression. It only requires a mean
square convergence rate for the regression learner $\hat{\gamma}_{k}$ and so
allows for a wide variety of first step machine learning estimators.

We could also formulate a non-asymptotic analog to this asymptotic result. This
would depend on the availability of non-asymptotic results for the learner
$\hat{\gamma}_{k}$. To the best of our knowledge such results are not
available for many learners, such as neural nets and random forests. To allow
the results of this section to include as many first steps as possible we
focus here on the asymptotic result and reserve the non-asymptotic result to
future work.

\section*{Acknowledgements}

The National Science Foundation provided partial financial support via
grants 1559172 and 1757140. Rahul Singh thanks the Jerry Hausman Dissertation Fellowship.


    \appendix

    \section{Related Work}\label{sec:related}

\subsection{Previous Learning Problems}

The paper builds upon ideas in classical semi- and nonparametric learning theory with low-dimensional $X$, using traditional smoothing methods [\cite{van1991differentiable}; \cite{newey1994asymptotic}; \cite{bickel1993efficient}; \cite{robins1995semiparametric}; \cite{vaart}], that do not apply to the current high-dimensional setting. Our paper also builds upon and contributes to the literature on modern orthogonal/debiased estimation and inference [\cite{zhang2014confidence}; \cite{belloni2011inference,belloni2014inference,belloni2015uniform}; \cite{javanmard2014confidence,javanmard2014hypothesis,javanmard2018debiasing}; \cite{van2014asymptotically}; \cite{ning2017general}; \cite{chernozhukov2015valid}; \cite{neykov2018unified}; \cite{ren2015asymptotic}; \cite{jankova2015confidence,jankova2016confidence,jankova2018semiparametric}; \cite{bradic2017uniform}; \cite{zhu2017breaking,zhu2018linear}], which focuses on coefficients in high-dimensional linear and generalized linear regression models, without considering the general linear functionals analyzed here.

The functionals we consider are different than those analyzed in \cite{cai2017confidence}. The continuity properties of functionals we consider provide additional structure that we exploit, namely the Riesz representer, an object that is not considered in \cite{cai2017confidence}. Targeted maximum likelihood, \cite{van2006targeted}, based on machine learners has been considered by \cite{van2011targeted} and large sample theory given by \cite{luedtke2016statistical}, \cite{toth2016tmle}, and Zheng et al. (2016). Here we provide DML learners via regularized RR, which are relatively simple to implement and analyze, and which directly target functionals of interest and learn the RR automatically from the data.

\subsection{De-biased Estimation}

We build on previous work on debiased estimating equations constructed by adding an influence function. \cite{hasminskii1979nonparametric} and \cite{bickel1988estimating} suggest such estimators for functionals of a density.
\cite{newey1994asymptotic} derives such scores as a part of the computation of the semi-parametric efficiency bound for regular functionals.  Doubly robust estimating equations as in \cite{robins1995analysis} and \cite{robins1995semiparametric} have this structure. \cite{newey1998undersmoothing,newey2004twicing} further develop theory in this vein, in a low-dimensional nonparametric setting. In the regular case, \cite{chernozhukov2016locally,chernozhukov2018double} analyze the double robust/debiased learners in several high-dimensional settings. However, analysis requires an explicit formula for the Riesz representer, used in its estimation, which is often unavailable in closed form (or may be inefficient when restrictions such as additivity are used---see Section~\ref{sec:examples} for the explicit definition of the additive model and structure of representers in that case). In contrast, here we estimate the Riesz representer \textit{automatically} from the moment conditions that characterize it, and extend the analysis to cover non-regular functionals.
   
Various papers have considered direct estimation of the Riesz representer. Among these papers, ours is the first to present a framework for direct estimation of the Riesz representer of a \textit{broad class of linear functionals, in a high-dimensional setting, without requiring strong Donsker class assumptions}. The earliest reference of which we know is \cite{robins2007comment}, a comment on another paper, which consider only the global average treatment effect (ATE). \cite{zhu2017breaking} show that it is possible to attain $\sqrt{n}$-consistency for the coefficients of a partially linear model when the regression function is dense. Our results apply to a much broader class of functionals, and allow for tradeoffs in accuracy of estimating the regression function and the Riesz representer. \cite{newey2018cross} present and analyze estimators based on regression splines, while we present and analyze sparse estimators methods for the high-dimensional setting. The \cite{athey2018approximate} estimator of the ATE is based on sparse linear regression and on approximate balancing weights when the regression is linear and strongly sparse. Our results apply to a much broader class of linear functionals and allow the regression learner to converge at relatively slow rates, including the dense case or approximately sparse case. 

Since the first version of this paper was posted online, subsequent work has built upon its insights. \cite{hirshberg2019augmented} build upon the present work by considering the problem of learning regular functionals when the regression function belongs to a Donsker class. They utilize the orthogonal representations proposed in this paper and \cite{chernozhukov2016locally}, and extend the initial version of the paper, \cite{hirshberg2017augmented}, that had only considered the ATE example. Our approach does not require a Donsker class assumption, which is too restrictive in our setting. \cite{hirshberg2018debiased} consider the average derivative functional in a single index model, analyzing a variant of the estimator proposed here, adapted to the single-index regression structure. \cite{rothenhausler2019incremental} builds upon our work, analyzing global average derivative functionals, and proposing practical Lasso-type solvers for estimating the RR. Our approach is also practical; the RR estimation is based on a Dantzig selector type estimator, which is easy to compute by linear programming methods. In follow-up work, \citet*{chernozhukov2018learning} consider different Lasso-type solvers for estimating RR.  Compared to \cite{rothenhausler2019incremental}, our analysis covers a much broader collection of functionals, and deals with both local and global versions. 

\subsection{Localized Functionals}

A new development incorporated in this version of the paper is the inclusion of local and localized functionals, such as average treatment/policy effects and derivatives localized to certain neighborhoods of a value of a low-dimensional covariate subvector. In low-dimensional nonparametrics, the study of such functionals, called ``partial means'' goes back, e.g., to \cite{newey1994kernel}.   In  contrast, here we treat the case where the ambient covariate space is very high-dimensional, but we localize with respect to a value of a low-dimensional subvector.  Moreover, we must rely on orthogonalized estimating equations to eliminate the regularization biases arising due to the high-dimensional ambient space. Various papers have studied debiased moment equations for certain localized functionals: conditional average treatment effect (CATE), continuous treatment effect (CTE), and regression derivative at a point. We instead present a unified analysis for the general class of localized functionals. Moreover, we cover local effects that are not perfectly localized, which may be more robust objects from an inferential point of view, as argued in \cite{genovese2008adaptive}.  

The debiased CATE and CTE literature is vast. Prominent examples of the debiased CATE literature include \cite{wang2010nonparametric}, \cite{van2014targeted}, \cite{luedtke2016statistical}, \cite{nie2017quasi}, \cite{lee2017doubly}, and most recently \cite{kennedy2020optimal}. Independently and contemporaneously to the present version of the paper,  \cite{lieli} and \cite{lechner} define and study perfectly localized average treatment effects with high-dimensional confounders. Prominent examples of the debiased CTE literature include \cite{rubin2006extending}, \cite{diaz2013targeted}, \cite{galvao2015uniformly}, \cite{kennedy2017nonparametric}, \cite{kallus2018policy}, and \cite{colangelo2020double}. These works develop inference on perfectly localized average potential outcomes with continuous treatment effects, using a different approach than what we develop here.  Our development is complementary as it covers a much broader collection of functionals.  

The debiased literature on regression derivative at a point is more recent. \cite{guo:zhang} study inference on the regression derivative $\partial \gamma_1(d)$ at a point $d$ in a high-dimensional regression model, $\gamma(D,Z) = \gamma_1(D) + \gamma_2(Z)$, where $D$ is univariate covariate of interest and $Z$ is a high-dimensional vector of control covariates.  Our analysis is again complementary: it covers objects like this, but also covers more general functionals like $\Ep [\partial_d \gamma(D,Z) \mid D= d]$, either without additivity structure or without requiring $D$ to be one-dimensional. \cite{semenova2020debiased}  apply low-dimensional series regression estimators on top of the pre-estimated unbiased orthogonal signal of treatment and partial derivative effects, where pre-estimation of the orthogonal signal is done in the high-dimensional setting. Our analysis has a rather different structure (without reliance on close-form solutions for Riesz representers), and kernels are used for localization instead of series.

Our work complements existing work that considers the problem of estimating general nonpathwise differentiable functionals like the localized ones here. Early contributions include \cite{robins2001comment}, \cite{van2003unified}, and \cite{rubin2005general}. More recently, \cite{athey2019generalized} consider this issue in the context of generalized random forests. \cite{foster2019orthogonal} present a general theory, but without inference guarantees. Unlike previous work, we analyze finite sample Gaussian approximation.

\section{Notation and preliminaries}\label{sec:preliminary}

\subsection{Notation glossary} 
 Let $W = (Y, X')'$ be a random vector with law $P$ on the sample space $\mathcal{W}$, and $W_1^n = (Y_i, X_i)_{i=1}^n$ denote i.i.d. copies of $W$.  The law of $X$ is denoted by $F$. All models and probability measure $P$ can be indexed by $n$, the sample size, so that the models and their dimensions and parameters determined by $P$ change with $n$. We use  notation from the empirical process theory, see \cite{van1996weak}. Let $\mathbb{E}_I f$ denote the empirical average of $f(W_i)$ over $i \in I \subset \{1,..., n\}$: $\mathbb{E}_I f  := \mathbb{E}_I f(W) = | I | ^{-1} \sum_{i \in I} f(W_i).$ Let $\Bbb{G}_I$ denote the empirical process over $f \in \mathcal{F}: \mathcal{W} \to \Bbb{R}^p$ and $ i \in I$, namely $\mathbb{G}_I f := \mathbb{G}_I f (W) := | I | ^{-1/2} \sum_{ i \in I} (f(W_i) - P f),$
where  $Pf :=  P f(W) := \int f(w) d P(w)$.  Denote the $L^q(P)$ norm of a measurable function $f: \mathcal{W} \to \Bbb{R}$ and also the $L^q(P)$ norm of random variable $f(W)$ by $\| f \|_{P,q} = \| f(W)\|_{P,q}$.  We use $\| \cdot\|_q$ to denote $\ell_q$ norm on $\Bbb{R}^d$.  For a differentiable map $x \mapsto f(x)$, from $\mathbb{R}^d$ to $\mathbb{R}^k$,  we use $\partial_{x'} f(x)$ to abbreviate the partial derivatives  $(\partial/\partial x') f(x)$, and we use 
$\partial_{x'} f(x_0)$ to mean $\partial_{x'} f (x) \mid_{x = x_0}$, etc.  We use $x'$ to denote the transpose of a column vector $x$. We say that $ a \lesssim b $ under the asymptotics with an index $n \to \infty$  if $a \leq C b $ for all $n$ sufficiently large, and  $a \asymp b$ if both $a \lesssim C b $ and $b \lesssim C a$ for all $n$ sufficiently large, where $C \geq 1$ is a positive constant that does not depend on $n$.

\subsection{Preliminaries}
To prove the first couple of lemmas we recall the following definitions and results. Given two normed vector spaces $V$ and $W$ over the field of  real numbers $\Bbb{R}$, a linear map $A : V \to W$ is continuous if and only if it has a bounded operator norm:
$$\displaystyle \|A\|_{op} :=\inf\{c\geq 0:\|Av\|\leq c\|v\|{\mbox{ for all }}v\in V\} < \infty,$$
where $\| \cdot \|_{op}$ is the operator norm. The operator norm depends on the choice of norms for the normed vector spaces $V$ and $W$.  A Hilbert space is a complete linear space equipped with an inner product $\langle f ,  g \rangle$
and the norm $ |\langle f,  f \rangle|^{1/2}$. The space $L^2(P)$ is the Hilbert space with the inner product
$\langle f, g \rangle = \int f g dP$ and norm $\|f\|_{P,2}$. The closed linear subspaces of $L^2(P)$ equipped with the same inner product and norm are Hilbert spaces.

\textbf{Hahn--Banach extension for normed vector spaces.}  If $V$ is a normed vector space with linear subspace $U$ (not necessarily closed) and if $\phi : U \mapsto K$ is continuous and linear, then there exists an extension $\psi: V \mapsto K$ of $\phi$ which is also continuous and linear and which has the same operator norm as $ \phi$.

\textbf{Riesz--Frechet representation theorem.} Let $H$ be a Hilbert space over $\Bbb{R}$ with an inner product $\langle \cdot , \cdot \rangle$, and $T$ a bounded linear functional mapping $H$ to $\Bbb{R}$. If $T$ is bounded then
there exists a unique $g \in H$ such that for every $f \in H$ we have $T(f) =\langle f, g\rangle$. It is given
by $g = z (T z)$, where $z$ is unit-norm element of the orthogonal complement of the kernel subspace
$K = \{ a \in H:  T a = 0\}$. Moreover, $\|T\|_{op} = \|g\|$,
where $\|T\|_{op}$ denotes the operator norm of $T$, while $\|g\|$ denotes the Hilbert space norm of $g$.

\textbf{Radon--Nykodym derivative.}  Consider a measure space $(\mathcal{X},\mathit\Sigma)$ 
on which two $\sigma$-finite measure  are defined, $\mu$ and $\nu$.
If $\nu \ll \mu$ (i.e. $\nu$ is absolutely continuous with respect to $\mu$), then there is a  measurable function $f: \mathcal{X} \rightarrow [0,\infty) $, such that for any measurable set $A \subseteq \mathcal{X} $,
$\nu(A) = \int_A f \, d\mu$. The function $f$ is conventionally denoted by $d\nu/d\mu$.

\textbf{Integration by parts.} Consider a closed measurable subset $\mathcal{X}$ of $\Bbb{R}^k$ equipped with Lebesgue measure $V$ and piecewise smooth boundary $\partial \mathcal{X}$, and suppose that $v: \mathcal{X} \to \Bbb{R}^k$ and 
$\phi: \mathcal{X} \to \Bbb{R}$ are both $C^1(\mathcal{X})$,
then 
$$
 {\displaystyle \int _{\mathcal{X} }\varphi \operatorname {div} { {v}}\,dV=\int _{\partial \mathcal{X} }\varphi \,{ {v}}' d{ {S}}-\int _{\mathcal{X} }{ {v}}' \operatorname {grad} \varphi \,dV,}
$$
where $S$ is the measure induced by $V$, and $n$ is outward-normal vector induced by $\mathcal{X}$. Here $\divg_d$ denotes the divergence
of a vector field $d \mapsto v(d)$: $\divg_d \ v = \langle \nabla, v \rangle = \sum_{j=1}^{\dim(d)} \partial_{d_j} v_j(d).$  

\section{Structure of functionals and their scores in leading examples}\label{sec:examples}

We see that the key quantities in the main inference results are  the operator norm $L$
of the linear functional and the standard deviation $\sigma$ and kurtosis $\kappa/\sigma$ of the score $\psi_0$. In this section we establish bounds on these quantities in the key Examples~\ref{ex:ate},~\ref{ex:policy1},~\ref{ex:policy2}, and~\ref{ex:deriv}, focusing on either unrestricted or additive nonparametric models. 

\subsection{Structure of Riesz representers for unrestricted and additive models}

Below we derive linear representers through change of measure and integration by parts.  
These representers are universal since they apply to the \textit{unrestricted model}, where $\bar{\Gamma} = L^2(F).$ We remark here that these representers are universal, since they can represent $\theta_0$ even when $\bar{\Gamma} \neq L^2(F)$,
if they exist.  These universal representers are not minimal unless $\bar{\Gamma} = L^2(F)$.  Theorem~\ref{theorem:eff} implies that it is better to use
the minimal representer than the universal representer to attain full semi-parametric efficiency (unless $\Gamma=L^2(F)$). 

Consider the following (some well-known) candidates  for universal linear representers in Examples~\ref{ex:ate},~\ref{ex:policy1},~\ref{ex:policy2}, and~\ref{ex:deriv}:
\begin{eqnarray}
\alpha_0(x; \ell) & =  &  \ [(1(d=1) - 1(d= 0) )/ \Pr(D=d \mid Z=z)] \bar \ell(z); \label{rr: ex1} \\
\alpha_0(x; \ell) & =  &  \ [d (F_1(x) - F_0(x))/d F(x)]  \ell(x); \label{rr: ex2}  \\
\alpha_0(x; \ell) & =  & \ [d (F_1(x) - F(x))/d F(x)] \ell(x), \   F_1 = \mathrm{Law} (T(X)) ; \label{rr: ex3} \\\
\alpha_0(x; \ell) & =  & - (\divg_d ( \ell(x) t(x) f(d|z) )/f(d|z), \  f(d|z) = \text{ pdf of } D \text{ given } Z=z \label{rr: ex4} ; 
\end{eqnarray}
for $\bar \ell(z) := \Ep[\ell(X)\mid Z=z]$;
treated as formal maps $\alpha_0: \mathcal{X} \to \Bbb{R} \cup \{ \mathrm{na} \}$,  where  $dF_k/dF$ denotes the Radon--Nykodym derivative of measure $F_k$ with respect to $F$ on $\mathrm{support}(\ell)$, and $\mathrm{na}$ is ``not available''.
The Radon--Nykodym derivatives exist if $F_k$ is absolutely continuous with respect to $F$ 
on $\mathrm{support}(\ell)$. 

\begin{lemma}[Universal representers for key examples]\label{lemma: key examples} In Examples~\ref{ex:ate},~\ref{ex:policy1},~\ref{ex:policy2}, and~\ref{ex:deriv},
(i) If $\alpha_0(X;\ell)$ is real-valued a.s. and $ \alpha_0(\cdot;\ell) \in L^2(F)$, then it is the universal representer for the corresponding linear functional $\gamma \mapsto \theta(\gamma)$, and the latter is continuous.   In Example~\ref{ex:deriv},  we require that $d \mapsto \gamma(x) \ell(x) t(x) f(d|z) $ is continuously differentiable on the support set $\mathcal{D}_z = \mathrm{support}(D|Z=z)$,  and vanishes on its boundary $\partial \mathcal{D}_z$, which is assumed to be piecewise-smooth, for each $ z \in \mathcal{Z}$.
Further, if $\bar{\Gamma} = L^2(F)$, the representer is minimal; otherwise, the minimal representer $\alpha^\star_0$ is obtained by projecting $\alpha_0$ onto $\bar{\Gamma}$.   (ii) There are examples of $P$, exhibited in the proof of this lemma, 
 such that linear functionals in Examples~\ref{ex:ate},~\ref{ex:policy1},~\ref{ex:policy2}, and~\ref{ex:deriv} can be continuous on $\bar{\Gamma} \neq  L^2(F)$, but  $\alpha_0(X;\ell) = \mathrm{na}$ with positive probability.\end{lemma}

Part of the lemma is well known (for example, the $\alpha_0(X;\ell)$ representer for ATE is the Horvitz-Thompson transformation), while a part of lemma appears to be new. The first part of the lemma provides a simple sufficient condition to guarantee continuity of the target functionals. 
It recovers well-known sufficient conditions for nonparametric identification of various functionals. 
The second part of the lemma states that this condition is not necessary, and that target functionals can be continuous on
some subsets of $L^2(F)$ without these conditions.

The following is a useful result in view of the wide practical use of additive models, which model
the regression function as additive in the two sets of vector components $x_1$ and $x_2$ of $x$.
(There is not much loss in generality in considering two sets rather than multiple sets).  It is an important setting
where $\Gamma$ is not dense in $L^2(F)$ and where minimal representers are not equal to the universal representers.

\begin{itemize}
\item[$\mathrm{AM}$] \textit{Suppose that the regression function is additive in components $x_1$ and $x_2$ of $x$:
$$x \mapsto \gamma(x) = \gamma_1(x_1) + \gamma (x_2), \quad x =(x_1', x_2')' \in \mathcal{X},$$ 
where $\gamma_1 \in \Gamma_{01}$,  a dense subset of $L^2(F_{1})$, where $F_1$ denotes the probability law of $X_1$. The linear functional $m_0$ and the weighing function $\ell$
depends only on the first component, namely $m (w, \gamma; \ell)= m(w, \gamma_1; \ell)$ and $\ell(x) = \ell(x_1)$. }
\end{itemize}

The following lemma shows that we can construct representers for additive models by taking
conditional expectation of a  universal representer.  We can immediately see that the minimal representers can be generated
as conditional expectations of the universal representers.

\begin{lemma}[Order-preserving, contractive representers for additive models]\label{lemma: additive}  Work with $AM$ and assume $\alpha_0 (\cdot; \ell) \in L^2(F)$.  Then on
$\gamma \in \Gamma$, 
$$
\theta(\gamma) = \theta(\gamma_1)= \int \alpha^\star_{0}(x_1) \gamma_1(x_1) dF(x_1), \ \ \alpha^\star_{0} (x_1) = \Ep [ \alpha_{0}(X) \mid X_1 = x_1],
$$
where $\alpha_{0}$ is any linear representer for $\gamma \mapsto \theta(\gamma)$ on $\Gamma$. In particular, the conditional expectation operator is order-preserving, and it induces the contraction for all $L^q(P)$ norms for all $q \in [1,\infty]$:
$$
\| \alpha^\star_0\|_{P,q} \leq \| \alpha_0 \|_{P,q}.
$$
\end{lemma}
The latter properties are useful in characterizing the structure of the global and local  functionals under condition AM.

\subsection{Structure of global functionals and scores in key examples}

Here we develop bounds on the key quantities:
the standard deviation $\sigma$ of the score, the kurtosis $\kappa/\sigma$, and the modulus of continuity $L$.
In the regular case, these quantities are bounded.  Here we would like to study
how the bounds depend on $L$, and we analyze the non-regular cases arising from 
taking a sequence of models with $L \to \infty$.

To make key points, we focus on the case  where either $\bar{\Gamma} = L^2(F)$ or $\bar{\Gamma} \subset L^2(F)$ 
with the additive model AM holding. Furthermore, we develop these bounds in the context of Examples~\ref{ex:ate},~\ref{ex:policy1}, and~\ref{ex:policy2}, though the proofs are useful to characterize bounds in other contexts.   Our goal is to fix a weighting function $\ell$,
and to consider how a non-regularity $L \to \infty $ can arise from modeling quantities like
\begin{equation}\label{eq: denoms}
1/\Pr(D=d \mid Z), \quad (d (F_1 -F_0)/dF)\circ X ,   \quad (d(F_1- F)/dF) \circ X,
\end{equation}
taking high values due to the denominator taking values close to zero. We may characterize
such cases as the weakening of overlap of supports of relevant distributions (e.g., $F$ puts small
mass on points where $F_1$ puts a lot of mass). In Example~\ref{ex:deriv},
a similar issue could arise due to $1/f(D|Z)$ taking high values; for brevity, we don't analyze this
source of non-regularity for Example~\ref{ex:deriv} and focus on localization as the source.

In the sequel, we say that $ a \lesssim b $ under the asymptotics with an index $n \to \infty$  if $a \leq C b $ for all $n$ sufficiently large, and  $a \asymp b$ if both $a \lesssim C b $ and $b \lesssim C a$ for all $n$ sufficiently large, where $C \geq 1$ is a positive constant that does not depend on $n$. 

\begin{lemma}[Structure of global average effects functionals and scores]\label{lemma: gstr13}  Suppose that either (a) $\bar{\Gamma} = L^2(F)$ or (b) that  $\bar{\Gamma} \subset L^2(F)$ 
with the additive model AM holding. Suppose that the universal Riesz representers $\alpha_0(X) = \alpha_0(X;\ell)$ given in formulae (\ref{rr: ex1}), (\ref{rr: ex2}), (\ref{rr: ex3})
for Examples~\ref{ex:ate},~\ref{ex:policy1}, and~\ref{ex:policy2} exist and are in $L^2(F)$. Suppose that $\alpha^\star_0(X) = \alpha_0(X)$ in the case (a) and $\alpha^\star_0(X_1) = \Ep [\alpha^\star_0 (X)\mid X_1]$ in the case (b) obey:
\begin{eqnarray}\label{switch norm}
 \|\alpha^\star_0\|_{P,3} \leq c ( \|\alpha^\star_0\|^2_{P,2} \vee 1),
 \end{eqnarray}
for some finite constant $c$  and that 
 $$U_1 =m(W,\gamma^\star_0(X))    - \Ep m(W,\gamma^\star_0(X)) \text{ and } U_2 = Y - \gamma_0^\star(X) $$
 obey the bounded moment and bounded heteroscedasticity conditions: 
 $$ (\Ep [|U_1|^q ] )^{1/q} \leq \bar c, \quad 0 < \underline c  \leq (\Ep [|U_2|^q |X] )^{1/q} \leq \bar c  \text{ a.s.,   for }  q \in \{2,3\},$$
 for some finite positive constants $\underline c$ and $\bar c$.  Then
$$
\underline c  L   \leq \sigma  \leq  \bar c  \sqrt{1+ L^2}, 
 \quad \kappa \leq  \bar c   (1+ c (L^2 \vee 1) ).
$$
If, as $n \to \infty$,  we have  that $L \to \infty$ and  the constants $(c,\underline c,\bar c)$ are bounded away from zero and above,  then
$$
(\kappa/\sigma) \lesssim \sigma \asymp L \to \infty.
$$
\end{lemma}

Condition (\ref{switch norm}) allows the $L^3(F)$ norm of the representer to be much larger than the $L^2(F)$ norm, but limits how much larger.  For instance, consider Example~\ref{ex:ate}. Suppose $\bar{\Gamma} = L^2(F)$ so that $\alpha^\star = \alpha_0$ and that the propensity score $P[D=1\mid Z]$ is uniformly distributed on $[\pi, 1/2]$. Then  $\| \alpha_0\|_{P,2} \asymp (1/\pi)^{1/2}$ and  $\| \alpha_0\|_{P,3} \asymp (1/\pi^2)^{1/3} \ll \| \alpha_0\|_{P,2} ^2$ when $\pi \searrow 0$, so the condition is easily met.

\subsection{Structure of local  and localized functionals and scores in key examples}

Here we focus on local functionals and develop bounds that relate key quantities: the standard deviation  $\sigma$ of the score, the kurtosis $\kappa/\sigma$, and the modulus of continuity $L$. 

Our first goal is examine how the
localization of the weighting function $\ell$ creates the non-regularity  $L \to \infty$.  Our inference theory outlined above covers  local functionals provided $L/\sqrt{n}$ is small,
and it also covers perfectly localized functionals provided the scaled localization bias is small:
$$
\sqrt{n} ( \theta (\gamma_0^\star; \ell_h)- \theta(\gamma_0^\star; \ell_0) )/\sigma  \to 0.
$$
We provide a bound on the localization bias in terms of the smoothness and the kernel order.
The latter additional requirement means that the inference on perfectly localized functionals is less robust than the inference on the local functionals (analogously, to the point that was made by \cite{genovese2008adaptive}).

\begin{lemma}[Structure of local average effects functionals and scores]\label{lemma: lstr13}  Suppose that either (a) $\bar{\Gamma} = L^2(F)$ or (b) $\bar{\Gamma} \subset L^2(F)$  with the additive model AM holding. Suppose the universal Riesz representer $\alpha_0(X;1)$, corresponding to the flat weighting function $\ell=1$,
given in formulae (\ref{rr: ex1}), (\ref{rr: ex2}), and (\ref{rr: ex3}), corresponding to Examples~\ref{ex:ate},~\ref{ex:policy1}, and~\ref{ex:policy2}, exists and obeys 
\begin{equation}\label{eq: shut down}
0< \underline{\alpha} \leq \alpha_0(X;1) \leq \bar \alpha, \quad \text{ a.s.}
\end{equation}
 Suppose for some $h_0>0$, we have that $N_{h_0}(d_0)= \{ d: \|d-d_0\|_\infty \leq h\} \subset \mathcal{D}$.
Suppose that for $\ell = \ell_h$ with $h  \leq h_0$:
$$U_1 =m(W, \gamma^\star_0(X); \ell )    - \Ep m(W,\gamma^\star_0(X); \ell ) \text{ and } U_2 = Y - \gamma_0^\star(X), $$
obey the bounded heteroscedastic moment conditions: 
 $$  (\Ep [|U_1|^q ] )^{1/q} \leq \bar c \| \ell\|_{P,q}, \quad 0 < \underline c  \leq (\Ep [|U_2|^q |X] )^{1/q} \leq \bar c  \text{ a.s.,   for }  q \in \{2,3\}.$$ 
Suppose that the pdf $f_D$ of $D$ obeys the bounds: $$ 0< \underline f \leq f_D(d) \leq \bar f \text{ and  }  \|\partial f_D(d)\|_1 \leq \bar f', \text{ for all } d \in N_{h_0}(d_0).$$  Then the non-asymptotic bounds stated in the proof of this lemma hold. In particular, if $h \searrow 0$ and  $(\underline \alpha, \bar \alpha, \underline c, \bar c, \underline f, \bar f, \bar f', h_0)$ are bounded away from zero and bounded above, then
$$
(\kappa/\sigma) \lesssim  h^{-p_1/6}  \lesssim \sigma \asymp L \asymp \| \ell\|_{P,2}  \asymp  h^{-p_1/2} \to \infty.
$$
\end{lemma}

The lemma shows that the main source of non-regularity is the bandwidth $h$ going to zero.  The condition (\ref{eq: shut down})
shuts down the previous source of non-regularity, and says that the quantities in (\ref{eq: denoms})
are now bounded from below and above. 

It is possible to analyze the case where both sources of non-regularity are present
and to bound behavior of $\sigma, \kappa/\sigma$, and $L$.
Our general inference theory allows for such complicated sources of nonregularity as long as these parameters
are much smaller than $\sqrt{n}$.

We now turn to characterization of the local average derivatives.

\begin{lemma}[Structure of local average derivative functionals and scores]\label{lemma: lstr4}  Suppose that either (a) $\bar{\Gamma} = L^2(F)$ or that  (b) $\bar{\Gamma} \subset L^2(F)$  with the additive model AM holding. 
Suppose the universal Riesz representer $\alpha_0(X; \ell_h)$ 
given in formula (\ref{rr: ex4}) exists for all $0< h< h_0$, where $h_0$ is a constant. Suppose that the errors  $$U_1 = m_0(W,\gamma^\star_0(X)) \ell_h(X) - \Ep m_0(W, \gamma^\star_0(X)) \ell_h(X) \text{ and } U_2 = Y - \gamma_0^\star(X)$$ obey
the bounded  heteroscedastic moment conditions:
$$ (\Ep [|U_1|^q] )^{1/q} \leq \bar c \| \ell_h\|_{P,q}, \quad  0 < \underline c  \leq (\Ep [|U_2|^q |X] )^{1/q} \leq \bar c, \text{ a.s.}, \quad q \in \{2,3\}.$$ 
Suppose  that $N_h(d_0)= \{ d: \|d-d_0\|_\infty \leq h\} \subset \mathcal{D}$ 
and that for all $d \in N_h(d_0)$: 
$$ 0< \underline f \leq f_D(d\mid Z) \leq \bar f , \quad  \|\partial f_D(d \mid Z)\|_1 \leq \bar f', \quad  t(d,Z) \leq \bar t, \quad |\mathrm{div_d} t(d,Z)| \leq \bar t'  \text{ a.s., }$$
$$ \Ep (t^2(d,X) | D=d ) \geq \underline t^2 \text{ for the case (a), } \Ep ( (\Ep[t (X) \mid X_1])^2 | D=d) \geq \underline t^2
\text{ for the case (b). }$$

Then the non-asymptotic bounds stated in the proof of this lemma hold.
In particular, if $h \searrow 0$ and $(\underline c, \bar c, \underline t, \overline t, \bar t', \underline f, \overline f, \bar f')$ are bounded away from zero and bounded above,  then
$$
\kappa/\sigma    \lesssim  h^{-p_1 /6}  \lesssim \sigma \asymp L \asymp h^{-p_1/2-1} \to \infty.$$

\end{lemma}

We next characterize the bias of approximating the perfectly localized parameter.  In what
follows the norm of a tensor $T =\partial^{\mathsf{v}}/(\partial d)^{\mathsf{v}}$ is defined
as the injective norm
$$
|T|_{op} = \sup_{\|u_1\|_2 \leq 1,..., \|u_\mathsf{v}\|_2 \leq 1 } |\langle T, u_1 \otimes .... \otimes u_\mathsf{v} \rangle |.
$$

\begin{lemma}[Structure of bias in perfect localization]\label{lemma: bias}  Suppose 
that for some $h_0>0$, $d \mapsto m(d)= \Ep [m(W, \gamma_0^\star) \mid D= d]$ and $d \mapsto f_D(d)$
 are continuously differentiable on $N_{h_0}(d_0)$ to the integer order $\mathsf{sm}$,  and for 
$\mathsf{v} := \mathsf{sm} \wedge \mathsf{o}$ and 
$\partial^\mathsf{v}_d$ denoting the tensor  $\partial^{\mathsf{v}}/(\partial d)^{\mathsf{v}}$ we have 
$$
\sup_{ d \in N_{h_0}(d_0)}  \| \partial^{\mathsf{v}}_d (m(d) f_D(d)) \|_{op} \leq \bar g_{\mathsf{v}}, \quad \sup_{ d \in N_{h_0}(d_0) } \|\partial^{\mathsf{v}}_d f_D(d)  \|_{op} \leq \bar f_{\mathsf{v}},
 \quad  \inf_{ d \in N_{h_0}(d_0) } f_D(d) \geq \underline{f}.
$$
In addition, assume
$$
m(d_0)f_D(d_0)\leq \bar{g}.
$$
We have that for all $h< h_1 \leq h_0$,
$$
|\theta(\gamma_0^\star; \ell_h)  - \theta(\gamma_0^\star; \ell_0)| \leq  C h^{ \mathsf{v}},$$
where the constant $C$ and $h_1$ depend  only on $K, \mathsf{v}, \bar g_{\mathsf{v}}$,  $\bar f_{\mathsf{v}}$, $\underline f, \bar{g}$.   If the latter constants
are bounded away from above and zero, as $h \searrow 0$, we have $|\theta(\gamma_0^\star; \ell_h)  - \theta(\gamma_0^\star; \ell_0)| \lesssim h^{ \mathsf{v}}.$
\end{lemma}

\section{Proofs for Section~2}

\subsection{Proof of Lemma~\ref{lemma: ERR}} We note that $\Gamma = \mathrm{span}(\Gamma_0)$ is a linear subspace of $L^2(F)$,
and $\bar \Gamma$ is a closed subspace by definition. Therefore, $\bar \Gamma$ is a Hilbert space with norm
$g \mapsto \| g\|_{P,2}$ and inner product $(f,g) \mapsto \langle f, g \rangle = \int f g dF$.

To show claim (i), we note that by the Hahn--Banach extension theorem, the operator $\theta: \Gamma \to \Bbb{R}$ can be extended 
to $\tilde \theta: \bar \Gamma \to \Bbb{R}$ such that $\| \tilde \theta\|_{op} = \|  \theta\|_{op}$. By the Riesz--Frechet theorem there exists
a unique representer $\alpha^\star_0$ such that $\tilde \theta (\gamma) = \langle \gamma, \alpha^\star_0 \rangle $ on $\gamma \in \bar \Gamma$ and $\| \tilde \theta\|_{op} = \| \alpha_0^\star \|_{P,2}$.

To show claim (ii), we are given a linear representer $\alpha_0$.
Denote by $\alpha^\star_0$ the projection of $\alpha_0$ onto $\bar \Gamma$.
Then
$\gamma \mapsto \varphi(\gamma) := \langle \gamma, \alpha_0 \rangle =\langle \gamma, \alpha^\star_0 \rangle$ agrees with $\gamma \mapsto \theta(\gamma)$ on $\gamma \in \Gamma$. Extend $\theta$ to $\bar \Gamma$
by defining $\tilde \theta(\gamma) = \varphi(\gamma) =  \langle \gamma, \alpha^\star_0 \rangle$ for $\gamma \in \bar \Gamma\setminus \Gamma$, which is well-defined by Cauchy-Schwarz inequality. Then $\| \varphi \|_{op} = \| \alpha^\star_0\|_{P,2} \leq  \| \alpha_0\|_{P,2} < \infty$, since the orthogonal projection reduces the norm. Further,
\begin{eqnarray*}
\infty>    \| \alpha^\star_0\|_{P,2}  & = &  \sup_{ \gamma \in \bar \Gamma\setminus\{0\}} | \langle \gamma, \alpha_0^\star \rangle|/ \| \gamma\|_{P,2} =  \sup_{\gamma \in \bar \Gamma\setminus\{0\}}  |\tilde \theta(\gamma) |/ \| \gamma\|_{P,2} = \| \tilde \theta\|_{op}.
\end{eqnarray*}
Hence $\alpha^\star_0$ is a representer for the extension $\tilde \theta$, and the Riesz--Frechet theorem implies that $\alpha^\star_0$  is unique.\qed

\section{Details for Section~3}\label{sec:diff}

\subsection{Practical implementation details} In practice we use the following generic algorithm for computing GDS estimators over subsamples $A$.
In particular, for regression we set $m(W, b) = Y b(X)$.
\begin{enumerate}
\item Obtain initial estimate $\hat t$ using a low-dimensional sub-dictionary $b_0$ of $b$:
$$
\hat t  \leftarrow (\hat t_0', 0')';\  \hat t_0 =  \hat G^{-1} \hat M_0;  \  \hat M_0 \leftarrow  \mathbb{E}_{A} m(W, b_0); 
 \  \hat G_0 \leftarrow  \mathbb{E}_{A} b_0 b_0';
$$
Compute the empirical moments for the full dictionary:
$$
\hat M  \leftarrow \mathbb{E}_{A} m(W, b);  \quad \hat G \leftarrow  \mathbb{E}_{A} b b'.
$$

\item  Update the diagonal normalization matrix:
$$\hat D^2 \leftarrow \mathrm{diag} \left( \mathbb{E}_{A} [ \{ b(X) b(X)' \hat t -  m(W, b)\}_j^2] ; \ \ j=1,...,p\right).$$

\item Update  the GDS estimate, using the current estimate as the starting point in the algorithm:
$$
\hat t \leftarrow \arg\min \| t\|_1 :  \|  \hat D^{-1} (\hat M  - \hat G t )\|_\infty \leq \lambda; \ \ \lambda = c \Phi^{-1}(1- \mathsf{a}/2p)/\sqrt{n}, 
$$

\item Iterate on steps 2 and 3 several times. Return the final estimate $ \hat t$.

\end{enumerate}

We note the following. First, theoretical arguments similar to \cite{belloni2012sparse} suggest that the data-driven algorithm behaves as the algorithm 
that knows the ideal $D$, since iterations yield $\|  D \hat D^{-1} -  I \|_\infty \to_{\Pr} 0 $.  The argument works provided we can set $c>1.1$ . In practice, however, $c=1$ works just fine from the outset.  We set $\mathsf{a}$ small, e.g. $\mathsf{a} =0.1$.

Second, \cite{chernozhukov2013gaussian} discuss finer data-driven choices of penalty levels based on the Gaussian or empirical bootstraps:
$$
\lambda = c \times [ (1-\alpha) -\mathrm{quantile} (  \|  \hat D^{-1} (\hat M^*  + \hat G^* t )\|_\infty \mid (W_i)_{i \in I^c_k})],
$$
where $\hat M^*$ and $\hat G^*$ are bootstrap copies of $\hat M$ and $\hat G$. This method yields an even lower theoretically valid penalty levels, because they adapt to the correlation structure much better. For instance, for highly-correlated empirical moments, the penalty level produced by this method can be substantially lower 
than the simple plug-in choice made above (in the extreme case, where the moments are perfectly correlated, the penalty level of \cite{chernozhukov2013gaussian} approximates $c\Phi^{-1}(1- \mathsf{a}/2))/\sqrt{n}$).


\subsection{Partial difference}

Consider a simplification of Example~\ref{ex:deriv}, average derivative:
$$
\theta^\star_0 = \int \partial_d \gamma_0^\star(d,z) \ell(x) dF(x).
$$
For nonparametric regression estimators that are linear in a dictionary $b(d,z)$, e.g. GDS and Lasso, the average derivative is straightforward to compute: apply the learned coefficients $\hat{\beta}$ to the derivative of the dictionary $\partial_d b(d,z)$, and average across observations using weighting $\ell(x)=\ell(d,z)$.

Random forest is an example of a nonparametric regression estimator that is not differentiable. A neural network is differentiable, but its derivative at each observation may be difficult to access when using a black-box implementation. For this reason, when using random forest or neural network, we use an average partial difference approximation of the average derivative.

Specifically, consider the average partial difference functional
$$
\theta_0^*=\int\left[\gamma_0^{\star}(d+\Delta/2,z)-\gamma_0^{\star}(d-\Delta/2,z)\right]\frac{1}{\Delta}\ell(x)dF(x).
$$
The theory developed for Example~\ref{ex:policy2}, policy effect from transporting $X$, directly applies to average partial difference. In practice, we take $\Delta$ to be one fourth of the standard deviation of $D$. 

There is an important connection between average derivative and average partial difference when using a nonparametric regression estimator that is linear in a dictionary $b(d,z)$, e.g. GDS and Lasso. If the dictionary $b(d,z)$ is quadratic in $d$, then the average derivative estimate must be \textit{numerically identical} to the average partial difference estimate. The specification from \cite{semenova2020debiased} that we use when estimating average price elasticity of gasoline is quadratic in log price. Therefore Table~\ref{table_APE} presents average partial difference estimates that perfectly coincide with average derivative estimates for GDS and Lasso, and that approximate average derivative estimates for random forest and neural network.

\subsection{Empirical results without debiasing}

We present tables analogous to those in Section~\ref{sec:applications} without debiasing. Tables~\ref{table_ATE_bias},~\ref{table_ATE2_bias}, and~\ref{table_APE_bias} in the supplement correspond to Tables~\ref{table_ATE},~\ref{table_ATE2}, and~\ref{table_APE} in the main text, respectively.

\begin{table}[h]
    \centering
    \caption{Average treatment effect of 401(k) eligibility on net financial assets without debiasing. Localized average
    treatment effects are reported by income quintile groups. The regression is estimated by GDS or Lasso. Standard errors are reported in parentheses.}
    \begin{tabular}{@{}ccccccc@{}}
    \hline 
        Income quintile & N treated & N untreated & \multicolumn{2}{c}{GDS}  &\multicolumn{2}{c}{Lasso}   \\
        \hline 
All & 3682 & 6187   &   3763.35   &   (31.01)& 4526.42   &   (42.33)\\
1 & 272 & 1702   &   2604.14   &   (8.05)&2581.88  &   (26.53)\\
2 & 527 & 1447   &    126.69   &   (5.92)&  298.56   &   (23.29)\\
3 & 755 & 1219   &    2819.64   &   (13.94)&2536.49   &   (28.56)\\
4  & 962 & 1012   &     5996.15  &   (57.05) & 3287.30   &   (84.56)\\
5   & 1166 & 807   &   4528.12   &   (103.84)& 6905.36   &   (159.28)\\
        \hline 
    \end{tabular}
    \label{table_ATE_bias}
\end{table}

\begin{table}[h]
    \centering
    \caption{Average treatment effect of 401(k) eligibility on net financial assets without debiasing. Localized average
    treatment effects are reported by income quintile groups. The regression is estimated by random forest or neural network. Standard errors are reported in parentheses.}
    \begin{tabular}{@{}ccccccc@{}}
    \hline 
        Income quintile & N treated & N untreated & \multicolumn{2}{c}{Random forest}&\multicolumn{2}{c}{Neural network} \\
        \hline 
All & 3682 & 6187   &   10543.48   &   (178.37)& 7807.97   &   (336.42)\\
1 & 272 & 1702   &   4378.26   &   (134.08)&4266.68  &   (308.06)\\
2 & 527 & 1447   &    1477.09   &   (329.52)&  1281.15  &   (537.07)\\
3 & 755 & 1219   &    6997.80  &   (158.49)&5331.58   &   (336.25)\\
4  & 962 & 1012   &    12854.02  &   (467.54) & 10234.88   &   (807.86)\\
5   & 1166 & 807   &   26845.23   &   (749.52)& 21426.42   &   (1615.20)\\
        \hline 

    \end{tabular}
    \label{table_ATE2_bias}
\end{table}

        

\begin{table}[h]
    \centering
     \caption{Estimated average derivative (price elasticity) of gasoline demand without debiasing. Localized average derivatives are
    reported by income quintile groups. The regression is estimated by GDS, Lasso, random forest, or neural network. Standard errors are reported in parentheses.}
    \begin{tabular}{@{}ccccccccccc@{}}
    \hline 
       Income quintile & N& \multicolumn{2}{c}{GDS} &\multicolumn{2}{c}{Lasso}&\multicolumn{2}{c}{Random forest} & \multicolumn{2}{c}{Neural network}\\
       \hline
        All   &  5001 & -0.53  &   (0.00)&-0.06  &   (0.00) & -0.09   &   (0.02)  &0.17   &   (0.01)\\
        1   & 1001& -0.55   &   (0.01)&0.00   &   (0.00)&-0.26   &   (0.07)  &0.18   &   (0.03)\\
        2   & 1000& -0.34   &   (0.01)&0.00   &   (0.00)&-0.15   &   (0.07)  &0.41   &   (0.03)\\
        3  & 1000 & -0.44   &   (0.01)&0.00   &   (0.00)&-0.30   &   (0.06)  &-0.21   &   (0.03)\\
        4   & 1000  & -0.22   &   (0.01)&0.00   &   (0.00) &-0.15   &   (0.07)  &0.23   &   (0.04)\\
         5  & 1000& -0.05   &   (0.00)&0.00   &   (0.00)&0.00   &   (0.07)  &0.61   &   (0.02)\\
        \hline 
        
    \end{tabular}
       \label{table_APE_bias}
\end{table}

\section{ Proofs for Section~4}

\subsection{Proof of Theorem~\ref{theorem: DML}}    The proof uses empirical process notation:  $\Bbb{G}_I$ denotes the empirical process over $f \in \mathcal{F}: \mathcal{W} \to \Bbb{R}^p$ and $ I \subset \{1,..., n\}$, namely $$\mathbb{G}_I f := \mathbb{G}_I f (W) := | I | ^{-1/2} \sum_{ i \in I} (f(W_i) - P f), \quad Pf :=  P f(W) := \int f(w) d P(w).$$

\textbf{Step 1.}  We have a random partition $(I_k,I^c_k)$
of $\{1,...,n\}$ into sets of size $m=n/K$ and $n- n/K$. Let
$$
\bar \theta_k = \theta_0 - \mathbb{E}_{I_k} \psi_0(W).
$$

Observe that in Lemma~\ref{lemma:score}, derivatives don't depend on $\theta$. Hence for all $\theta$,
$$
\partial_\beta \psi (W,  \theta; \beta_0, \rho_0) = -m(W, b) + {\rho_0}' b(X) b(X) =:  \partial_\beta \psi_0 (W) 
 $$ $$
 \partial_\rho \psi (W,  \theta; \beta_0, \rho_0) =  -b(X) (Y - b(X)'\beta_0) =:  \partial_\rho \psi_0 (W)
 $$ $$\partial^2_{\beta \rho'} \psi (X,  \theta; \beta_0, \rho_0) = b(X) b(X)' =: \partial^2_{\beta \rho'}  \psi_0 (W),
$$
where $\psi_0(W) :=\psi(W, \theta_0; \beta_0, \rho_0)$ as before.

Define the estimation errors $u := \hat \beta_{k} - \beta_0$ and $v:= \hat \rho_{k} - \rho_0.$
Using Lemma~\ref{lemma:score}, we have by the exact Taylor expansion around $(\beta_0, \rho_0)$
$$
\hat \theta_k =  \bar \theta_k -  (\mathbb{E}_{I_k}  \partial_{\beta} \psi_0(W))' u -
(\mathbb{E}_{I_k} \partial_{\rho} \psi_0(W))' v - u'(\mathbb{E}_{I_k} \partial^2_{\beta \rho'}  \psi_0(W)) v.
$$
Consider the event $\mathcal{E}$ that Condition R holds.  On this event:
\begin{eqnarray*}
(\sqrt{m}/\sigma) (\hat \theta_k - \bar \theta_k) & = & \mathrm{rem}_k   :=  
 \sum_{j=1}^4 \mathrm{rem}_{jk} :=      -\sigma^{-1} [\mathbb{G}_{I_k}  \partial_{\beta} \psi_0(W)] 'u -    \sigma^{-1} [\mathbb{G}_{I_k} \partial_{\rho} \psi_0(W)]' v   
\\  && -   \sigma^{-1} u ' [\mathbb{G}_{I_k}  \partial^2_{\beta \rho'}\psi_0(W)]v 
-      \sigma^{-1} \sqrt{m}  u' [P \partial^2_{\beta \rho'}  \psi_0(W)] v, 
\end{eqnarray*}
where we have used that by Lemma~\ref{lemma:score} 
$$
P \partial_{\beta} \psi_0(W)'u    =0, \ \
P \partial_{\rho} \psi_0(W)'v   =0.
$$

We now bound $\Ep[\mathrm{rem}_k^21(\mathcal{E})]$ by analyzing each of its terms. By the law of iterated expectations
\begin{align*}
    \Ep[\mathrm{rem}_k^21(\mathcal{E})]
    &=\Ep[\Ep[\mathrm{rem}_k^21(\mathcal{E})|(W_i)_{i \in I^c_k}]] 
    \leq 4 \sum_{j=1}^4 \Ep[\Ep[\mathrm{rem}_{jk}^21(\mathcal{E})|(W_i)_{i \in I^c_k}]] 
\end{align*}
using the fact that
$ \Ep \left(\sum_{j=1}^J V_j\right)^2 \leq J \sum_{j=1}^J \Ep V^2_j$
for arbitrary random variables $(V_j)_{j=1}^{J}$.

Note that $u$ and $v$ are fixed once we condition on the observations $(W_i)_{i \in I^c_k}$. 
On the event $\mathcal{E}$, by condition R, $\rem_{1k}, \rem_{2k}$ and $\rem_{3k}$ have conditional mean 0 and conditional variance given by
\begin{eqnarray*}
 \sigma^{-1} \sqrt{Var} [ \rem_{1k} \mid (W_i)_{i \in I^c_k}]
  & = & \sigma^{-1} \sqrt{Var}  [(\partial_{\beta} \psi_0(W)' u) \mid (W_i)_{i \in I^c_k}] \\
 & \leq & \sigma^{-1} \mu \sigma \sqrt{u' G  u} = \sigma^{-1} \mu \sigma r_1 \leq \delta , \\
\sigma^{-1}\sqrt{Var} [  \rem_{2k}  \mid (W_i)_{i \in I^c_k}]
 & = &   \sigma^{-1} \sqrt{Var} [ (\partial_{\rho} \psi_0(W) ' v)\mid (W_i)_{i \in I^c_k}]\\
 &\leq & \sigma^{-1}\mu \sqrt{v' G v} = \sigma^{-1} \mu \sigma r_2 \leq \delta, \\
\sigma^{-1}\sqrt{Var}  [ \rem_{3k} \mid (W_i)_{i \in I^c_k}]
& = &  
\sigma^{-1}\sqrt{Var}  [ u '  b(X) b(X)' v \mid (W_i)_{i \in I^c_k}]\\
 & \leq &  \sigma^{-1} \mu ( \sqrt{v' G v} + \sqrt{u' G u})  \\
& \leq &   \sigma^{-1} \mu (\sigma r_2 + r_1)     \leq \delta.
\end{eqnarray*}
On the event $\mathcal{E}$, $\mathrm{rem}_{4k}$ has conditional mean and conditional variance given by
\begin{align*}
    |\sigma^{-1}\sqrt{m}  u ' [P \partial^2_{\beta \rho'}  \psi_0(W)] v|
\leq  \sigma^{-1}\sqrt{m} \sigma r_3 
\leq \delta,  \sqrt{Var} [\rem_{4k} \mid (W_i)_{i \in I^c_k}]
=
0.
\end{align*}

In summary,
$$
 \Ep[\mathrm{rem}_k^21(\mathcal{E})]\leq 4 [\delta^2+\delta^2+\delta^2+\delta^2]=16\delta^2.
$$

\textbf{Step 2.} Here we bound the difference between $\hat \theta = K^{-1} \sum_{k=1}^K \hat \theta_{k}$
and $\bar \theta = K^{-1} \sum_{k=1}^K \bar \theta_{k}$:
$$
\sqrt{n}/\sigma|\hat \theta - \bar \theta| \leq \frac{\sqrt{n}}{\sqrt{m}} \frac{1}{K} \sum_{k=1}^K  \sqrt{m/\sigma} | \hat \theta_{k} - \bar \theta_{k}| \leq \frac{\sqrt{n}}{\sqrt{m}}  \frac{1}{K} \sum_{k=1}^K \mathrm{rem}_{k}.
$$
By Markov inequality we have 
\begin{eqnarray*}
&& \Pr\left(  \frac{1}{K}    \sum_{k=1}^K \mathrm{rem}_{k}> 4 \delta/\Delta\right)   \leq   \Pr \left ( \frac{1}{K}    \sum_{k=1}^K \mathrm{rem}_{k} > 4 \delta/\Delta \cap \mathcal{E} \right ) + \Pr\left (\mathcal{E}^c \right )\\
&&   \leq K^{-2} \Ep  \left ( \left(\sum_{k=1}^K \mathrm{rem}_{k}\right)^2 1( \mathcal{E}) \right )  \Delta^2 /(16\delta^2) + \epsilon  \\
 && \leq  K^{-2} K^2 \max_k  \Ep (\mathrm{rem}^2_k 1( \mathcal{E}))  \Delta^2 /(16\delta^2) + \epsilon \leq   \Delta^2 + \epsilon.
\end{eqnarray*}
And we have that $\sqrt{n/m} = \sqrt{ K}$. So it follows that
$$
 |\sqrt{n}(\hat \theta - \bar \theta)/\sigma| \leq   \mathsf{err}= 4 \sqrt{K} \delta/\Delta
$$
with probability at least $1-  \Pi$ for $\Pi:= \Delta^2 + \epsilon$.

\textbf{Step 3}. To show the second claim, let $Z :=  \sqrt{n}(\bar \theta - \theta_0)/\sigma$. By the Berry--Esseen bound, for some absolute constant $A$,
$$
\sup_{z \in \Bbb{R}} |\Pr ( Z \leq z) - \Phi(z)| \leq A \| \psi_0/\sigma \|^3_{P,3} n^{-1/2} = A (\kappa/\sigma)^{3} n^{-1/2}.
$$
The current best estimate of $A$ is 0.4748, due to \cite{shevtsova2011absolute}.
Hence, using  Step 2, for any $z \in \Bbb{R}$, we have
\begin{eqnarray*}
    &&  \Pr( \sqrt{n}(\hat \theta - \theta_0)/\sigma  \leq z) - \Phi(z) =\Pr( \sqrt{n} (\hat \theta - \bar\theta)/\sigma +Z \leq z)  - \Phi(z)\\
    && =\Pr( Z \leq z+\sqrt{n}(\bar\theta-\hat \theta)/\sigma )  - \Phi(z) \leq \Pr( Z \leq z+\mathsf{err})+\Pi - \Phi(z) \\
    && =\Pr( Z \leq z+\mathsf{err})-\Phi(z+\mathsf{err})+ \Phi(z+\mathsf{err}) - \Phi(z)+\Pi  \\
    && \leq A  (\kappa/\sigma)^{3} n^{-1/2} +   \mathsf{err}/ \sqrt{2 \pi}  + \Pi,
\end{eqnarray*}
where $1/\sqrt{2 \pi}$ is the upper bound on the derivative of $\Phi$.
Similarly,  conclude
that  $$\Pr ( \sqrt{n} \sigma^{-1} (\hat \theta - \theta_0) \leq z) -\Phi(z) \geq   A  (\kappa/\sigma)^{3} n^{-1/2} - \mathsf{err}/ \sqrt{2 \pi}  - \Pi.$$
The result follows by noting that $4/\sqrt{2 \pi} = 1.5957... < 2.$
\qed

\subsection{Proof of Theorem~\ref{theorem:eff}} We shall verify the hypotheses of \cite{vaart}, Theorem~25.20.

\textbf{Step 1.} Suppose that
$W$ had Radon--Nykodym derivative $dP$ under $P$ with respect to some measure $\mu$. 
Consider the set for some $\varepsilon>0$:
$$
\mathcal{S}_\varepsilon = \{ \delta \text{ measurable } : \mathcal{W}  \to \Bbb{R},  \int \delta dP  =0,\, \|\delta
\|_\infty \leq 1/(2\varepsilon)\}.
$$
Consider a
parametric submodel (i.e. path) of the form%
\[
\mathcal{P} = \Big \{
dP_{\tau}\left(  w\right)  =dP\left(  w\right)  \left[  1+\tau\delta\left(
w\right)  \right]  : \ \ \delta \in \mathcal{S}_\varepsilon \}_{\tau \in (0,\varepsilon)}.
\]
It is standard to verify that $\delta$ is the score of $dP_\tau$, namely $\delta(w) = \partial_\tau \log dP_\tau (w)$, and that
quadratic mean differentiability holds:
$$
\int [  (\sqrt{dP}_\tau - \sqrt{dP})/\tau - (\delta/2) d \sqrt{dP}]^2 d\mu \to 0,
$$
which implies that deviations from $P$ are locally asymptotically normal.  
The collection of scores $\mathcal{S}_\varepsilon$ therefore form the tangent set of $\mathcal{P}$ at $P$.

Consider the parameter of interest:
$$
\theta_\tau = \int m(w, \gamma_{\tau}) dP_\tau,
$$
where $\gamma^\star_{\tau}$ abbreviates the heavy notation $\gamma^\star_{0, P_\tau}$, denoting  the projection of $Y$ on $\bar \Gamma$ under $P_\tau$.  We will also use  $\gamma^\star_{0}$ to denote  $\gamma^\star_{0, P}$.

Step 2 below shows the differentiability of the parameter with respect to $\tau$:
$$
\frac{\theta_\tau-  \theta_0}{\tau} \to \int \psi_0 \delta dP,  \text{ for each }  \delta \in \mathcal{S}_\varepsilon,
$$
where $\psi_0$ is a score function. This is done in Step 2 below.  

This score function belongs to the $L^2(P)$ closure of the linear span of $\mathcal{S}_\varepsilon$:
$$
\overline{\mathrm{span}} (\mathcal{S}_\varepsilon) = \Big \{ \delta \in L^2(P): \int \delta dP = 0 \Big\}.$$
so it follows that $\psi_0$ is the projection of itself on the $\mathcal{S}_\varepsilon$ and is therefore the
only influence function.

\textbf{Step 2}.  Because $\delta$ is bounded by $1/(2 \varepsilon)$, the  $dP_{\tau}$ and
$dP$ dominate each other so that $\bar \Gamma$ does not depend on
$\tau$.  Let $\Ep_{\tau}$ denote expectation
under $P_\tau$ and $\Ep$ under $P$.

Then for some generic positive
finite constant $C$
\[
\Ep\left[ \gamma^\star_{\tau}\left(  X\right)  ^{2}\right]  \leq C\Ep_{\tau}\left[
\gamma^\star_{\tau}\left(  X\right)  ^{2}\right]  \leq C\Ep_{\tau}\left[
Y^{2}\right]  \leq C\Ep\left[ Y^{2}\right]  =C.
\]
Note that by $\gamma^\star_{\tau},\gamma^\star_{0}\in \bar \Gamma$ and the previous inequality, as
$\tau\rightarrow0$%
$$
\Ep\left[ \gamma^\star_{\tau}\left(  X\right)  \gamma^\star_{0}\left(  X\right)  \right]
=\Ep_{\tau}\left[  \gamma^\star_{\tau}\left(  X\right)  \gamma^\star_{0}\left(  X\right)
\right]  +o\left(  1\right) $$
$$ =\Ep_{\tau}\left[  Y\gamma^\star_{0}\left(  X\right)
\right]  +o\left(  1\right)  =\Ep\left[ Y\gamma^\star_{0}\left(  X\right)  \right]
+o\left(  1\right)  =\Ep\left[ \gamma^\star_{0}\left(  X\right)  ^{2}\right]
+o\left(  1\right)  .
$$
Similarly we have%
$$
\Ep\left[ \gamma^\star_{\tau}\left(  X\right)  ^{2}\right]    =\Ep_{\tau}\left[
\gamma^\star_{\tau}\left(  X\right)  ^{2}\right]  +o\left(  1\right)  =\Ep_{\tau
}\left[  Y\gamma^\star_{\tau}\left(  X\right)  \right]  +o\left(  1\right)$$$$
=\Ep\left[ Y\gamma^\star_{\tau}\left(  X\right)  \right]  +o\left(  1\right)
=\Ep\left[ \gamma^\star_{0}\left(  X\right)  \gamma^\star_{\tau}(X)\right]  +o\left(
1\right)   \rightarrow \Ep[\gamma^\star_{0}(X)^{2}].
$$
Therefore it follows that%
$$
\Ep\left[ \left\{  \gamma^\star_{\tau}\left(  X\right)  -\gamma^\star_{0}\left(  X\right)
\right\}  ^{2}\right]  =\Ep\left[ \gamma^\star_{\tau}\left(  X\right)  ^{2}\right]
+\Ep\left[ \gamma^\star_{0}\left(  X\right)  ^{2}\right]  -2\Ep\left[ \gamma^\star_{\tau
}\left(  X\right)  \gamma^\star_{0}\left(  X\right)  \right]  \rightarrow0.
$$
Note that $\left\vert \Ep\left[ \alpha_{0}\left(  X\right)  \left\{
\gamma^\star_{\tau}\left(  X\right)  -\gamma^\star\left(  X\right)  \right\}
\delta\left(  W\right)  \right]  \right\vert \leq C\Ep\left[ \left\vert
\alpha_{0}\left(  X\right)  \right\vert \left\vert \gamma^\star_{\tau}\left(
X\right)  -\gamma^\star_{0}\left(  X\right)  \right\vert \right]  \rightarrow0$
so that
\begin{align*}
\Ep\left[ m\left(  W,\gamma^\star_{\tau}\right)  \right]  -\Ep\left[ m\left(
W,\gamma^\star_{0}\right)  \right]    & =\Ep\left[ \alpha_{0}\left(  X\right)
\left\{  \gamma^\star_{\tau}\left(  X\right)  -\gamma^\star_{0}\left(  X\right)  \right\}
\right]  \\
& =\Ep_{\tau}\left[  \alpha_{0}\left(  X\right)  \left\{  \gamma^\star_{\tau}\left(
X\right)  -\gamma^\star_{0}\left(  X\right)  \right\}  \right]  \\
& -\tau \Ep \left[
\alpha_0 \left(  X\right)  \left\{  \gamma^\star_{\tau}\left(  X\right)  -\gamma^\star
_{0}\left(  X\right)  \right\}  \delta\left(  W\right)  \right]  \\
& =\Ep_{\tau}\left[  \alpha_{0}\left(  X\right)  \left\{  Y-\gamma^\star_{0}\left(
X\right)  \right\}  \right]  +o\left(  \tau\right)  \\
& =\Ep_{\tau}\left[  \alpha_{0}\left(  X\right)  \left\{  Y-\gamma^\star_{0}\left(
X\right)  \right\}  \right]  -\Ep\left[ \alpha_{0}\left(  X\right)  \left\{
Y-\gamma^\star_{0}\left(  X\right)  \right\}  \right]  +o\left(  \tau\right)  \\
& =\tau \Ep\left[ \alpha_{0}\left(  X\right)  \left\{  Y-\gamma^\star_{0}\left(
X\right)  \right\}  \delta\left(  W\right)  \right]  +o\left(  \tau\right)  .
\end{align*}
Therefore $\Ep\left[ m\left(  W,\gamma^\star_{\tau}\right)  \right]  $ is
differentiable at $\tau=0$ with $$\partial \Ep\left[ m\left(  W,\gamma^\star_{\tau
}\right)  \right]  /\partial\tau=\Ep\left[ \alpha_{0}\left(  X\right)  \left\{
Y-\gamma^\star_{0}\left(  X\right)  \right\}  \delta\left(  W\right)  \right].$$ In
addition, by mean-square continuity of $m\left(  W,\gamma^\star\right)  $,
\begin{align*}
\Ep_{\tau}\left[  m\left(  W,\gamma^\star_{\tau}\right)  \right]  -\Ep[m(W,\gamma^\star_{\tau
})]  & =\tau \Ep\left[ m\left(  W,\gamma^\star_{\tau}\right)  \delta(W)\right]  \\
&=\tau
\Ep\left[ m\left(  W,\gamma^\star_{0}\right)  \delta(W)\right]  +\tau \Ep[\{m(W,\gamma^\star
_{\tau})-m(W,\gamma^\star_{0})\}\delta(W)]\\
& =\tau \Ep\left[ m\left(  W,\gamma^\star_{0}  \right)  \delta(W)\right]  +o\left(
\tau\right)  .
\end{align*}
It follows that $\Ep_{\tau}\left[  m\left(  W,\gamma^\star_{\tau}\right)  \right]
-\Ep[m(W,\gamma^\star_{\tau})]$ is differentiable with%
\[
\frac{\partial\{\Ep_{\tau}\left[  m\left(  W,\gamma^\star_{\tau}\right)  \right]
-\Ep[m(W,\gamma^\star_{\tau})]\}}{\partial\tau}=\Ep\left[ m\left(  W,\gamma^\star_{0}\right)
\delta(W)\right]  =\Ep[\{m\left(  W,\gamma^\star_{0}\right)  -\theta_{0}\}\delta(W)].
\]
It then follows by the derivative of the sum being the sum of the derivatives
that $\theta_{\tau}=\Ep_{\tau}[m(W,\gamma^\star_{\tau})]$ is differentiable at $\tau=0$
and
\[
\frac{\partial\theta_{\tau}}{\partial\tau}=\Ep[\psi_0(W)\delta(W)].
\]
\qed

\subsection{Proof of Lemma~\ref{lemma: boundS}} First, we note that $$\|t_0^{\mathcal{M}}\|_0 = | \mathcal{M}| \leq s:=\max\{ x:  A x^{-a} \geq \nu \} = (A/\nu)^{1/a}.$$
Define
$$
t^r := t_0 - t_0^{\mathcal{M}} = t_0 1( |t_0| \leq \nu).
$$
Note that  $$\| t^r \|_1 \leq \nu s +  \int_{s}^\infty A x^{-a} dx  =  \nu s - \frac{1}{1-a} A s^{-a+1}= \nu s - \frac{1}{1-a} \nu s = \frac{a}{a-1} \nu s.$$

Then $\delta \in  S (t_0, \nu)$ implies that, by the repeated use of the triangle inequality:
$$
\| t_0 + \delta\|_1 \leq \| t_0\|_1 \Longleftrightarrow    \| t^{\mathcal{M}}_0 + \delta_{\mathcal{M}}\|_1 + \| t_0^r + \delta_{\mathcal{M}^c} \|_1   \leq \|t_0^{\mathcal{M}}\|_1 + \|t^r_0\|_1 
$$$$
\implies    \| \delta_{\mathcal{M}^c} \|_1 -\|t_0^r\|_1   \leq \| t_0^r + \delta_{\mathcal{M}^c} \|_1   \leq \|t_0^{\mathcal{M}}\|_1 -  \| t^{\mathcal{M}}_0 + \delta_{\mathcal{M}}\|_1 + \|t^r_0\|_1 
$$$$
\implies \| \delta_{\mathcal{M}^c}\|_1 -\| t^r_0 \|_1  \leq \| \delta_{\mathcal{M}}\|_1 + \| t^r_0 \|_1 \implies \| \delta_{\mathcal{M}^c}\|_1 \leq \| \delta_{\mathcal{M}}\|_1 + 2 \| t^r_0 \|_1.
$$

If $2 \| t^r \|_1\leq \| \delta_{\mathcal{M}}\|_1$, we have that $\| \delta_{\mathcal{M}^c}\|_1 \leq 2\| \delta_{\mathcal{M}}\|_1$, so using
the definition of the cone invertibility factor we obtain
$$
(k/s) \|\delta\|_1 \leq  \|G\delta\|_\infty \leq \nu \implies  \delta'G \delta \leq \| \delta\|_1 \| G \delta\|_\infty \leq (s/k) \nu^2.
$$

If $2 \| t^r \|_1\geq \| \delta_{\mathcal{M}}\|_1$, then $\|\delta\|_1 \leq 6 \| t^r \|_1$
$$
\delta'G \delta \leq \| \delta\|_1 \| G \delta\|_\infty \leq  6 \| t^r \|_1 \nu \leq 6 \frac{a}{a-1} s \nu^2.  \quad \qed
$$

\subsection{Proof of Lemma~\ref{lemma:RMD}}  Consider the event $\mathcal{R}$ such that 
\begin{equation}\label{En1}  \quad \| \hat g(t_0)\|_\infty \leq  \lambda,  \ \ \quad \| \hat g(\hat t)\|_\infty \leq \lambda,
\end{equation} holds.  This event holds with probability at least $1- \epsilon$.
The event  $\mathcal{R}$ implies that $\|\hat t\|_1 \leq \| t_0 \|_1$ by definition of $\hat t$, which further implies that
for  $ \delta = \hat t - t_0$
\begin{eqnarray*}
 \| G \delta \|_\infty & \leq &  \|  (G- \hat G) \delta \|_\infty +  \| \hat G \delta \|_\infty\\
 & = &   \|  (G- \hat G) \delta \|_\infty + \| \hat g(\hat t)- \hat g(t_0)\|_\infty \\
& \leq &   \|  G- \hat G\|_\infty \|\delta \|_1 + \| \hat g(\hat t)\|_\infty + \| \hat g(t_0)\|_\infty \\
& \leq & \bar \lambda 2B  +  2 \lambda \leq  \bar \nu .
\end{eqnarray*}
Hence $\delta \in S (t_0,\nu)$ with probability $1- \epsilon$.

The first inequality now in the bound follows from the definition of $s(t_0)$:
$
\sup_{\delta \in S (t_0,\nu)} \delta'G \delta \leq s(t_0) \nu^2.
$
The second bound follows by $\|\delta\|_1 \leq 2B$,
$
 \delta' G \delta \leq  \| G \delta \|_\infty \|\delta\|_1 \leq   \nu 2 B. $ \qed

\subsection{Proof of Theorem~\ref{theorem: rates1} and Corollary \ref{Cor:SUFF}} Application of Lemma~\ref{lemma:RMD} implies that with probability at least $1-4\epsilon$,
estimation errors $\tilde u = D^{-1}_\beta (\hat \beta_A - \beta_0)$ and $\tilde v = D^{-1}_\rho (\hat \rho_A - \rho_0)$ obey
$$
\tilde u'G \tilde u \leq C  [ (B^2  \tilde \ell^2 s (D^{-1}_\beta  \beta_0; \nu ) / n)  \wedge  ( B^2 \tilde \ell /\sqrt{n} )], $$$$
 \tilde v'G\tilde v  \leq C [  ( B^2 \tilde \ell^2  s (D^{-1}_\rho \rho_0; \nu ) / n)  \wedge  ( B^2 \tilde \ell /\sqrt{n)}],$$
 where $C$ is an absolute constant. Then
 $$
  |u'G u | \leq \mu_D^2  \tilde u'G \tilde u, \quad   |v'Gv| \leq \mu_D^2  \sigma^2   \tilde v'G\tilde v.
 $$
The stated bounds then follow. Hence the guarantee $R(\delta)$ holds for $\varepsilon = 1 - K 4 \epsilon$ provided that
 for some large enough absolute $C$:
$$
 C \sigma^{-1} (\sqrt{m} \sigma r_3  +  \mu r_1 ( 1+ \sigma)  + \mu \sigma r_2 )  \leq \delta,
$$
for $r_1$, $r_2$, and $r_3$ given in the corollary.  \qed

\section{Proofs for Section~5}

\subsection{Proof of Theorem~\ref{theorem: asymptotic}}
Let $\phi(w,\gamma,\alpha)=\alpha(x)[y-\gamma(x)],$
$\psi(w,\gamma,\alpha,\theta)=\theta-m(w,\gamma)-\phi(w,\gamma,\alpha),$
$\bar{\phi}(\gamma,\alpha)=\int\phi(w,\gamma,\alpha)F_{0}(dw),$ and $\bar
{m}(\gamma)=\int m(w,\gamma)F_{0}(dw).$ Note that%
\begin{equation}
\bar{\phi}(\gamma^{\star}_{0},\alpha_{0}^{\star})=0\text{, }\bar{\phi}(\gamma_{0}%
^{\star},\hat{\alpha}_{k})=0,\text{ }\bar{m}(\hat{\gamma}_{k}-\gamma_{0}^{\star
})=-\bar{\phi}(\hat{\gamma}_{k},\alpha_{0}^{\star}). \label{doubly robust}%
\end{equation}
Then we have
\begin{align*}
&\hat{\theta}_{k}  -\theta_{0}+\frac{1}{n_{k}}\sum_{i\in I_{k}}\psi
_{0}^{\star}(W_{i})=\frac{1}{n_{k}}\sum_{i\in I_{k}}\{\psi(W_{i},\gamma
_{0}^{\star},\alpha_{0}^{\star},\theta_{0})-\psi(W_{i},\hat{\gamma}_{k}%
,\hat{\alpha}_{k},\theta_{0})\}\\
&  =\frac{1}{n_{k}}\sum_{i\in I_{k}}\{m(W_{i},\hat{\gamma}_{k})+\phi
(W_{i},\hat{\gamma}_{k},\hat{\alpha}_{k})-m(W_{i},\gamma_{0}^{\star}%
)-\phi(W_{i},\gamma_{0}^{\star},\alpha_{0}^{\star})\}=\hat{R}_{1}+\hat{R}_{2},
\end{align*}
where%
\begin{align}
\hat{R}_{1}  &  =\frac{1}{n_{k}}\sum_{i\in I_{k}}[m(W_{i},\hat{\gamma}%
_{k}-\gamma_{0}^{\star})-\bar{m}(\hat{\gamma}_{k}-\gamma_{0}^{\star
})]\label{remain}\\
&  +\frac{1}{n_{k}}\sum_{i\in I_{k}}[\phi(W_{i},\hat{\gamma}_{k},\alpha
_{0}^{\star})-\phi(W_{i},\gamma_{0}^{\star},\alpha_{0}^{\star})-\bar{\phi}%
(\hat{\gamma}_{k},\alpha_{0}^{\star})]\nonumber\\
&  +\frac{1}{n_{k}}\sum_{i\in I_{k}}[\phi(W_{i},\gamma_{0}^{\star},\hat{\alpha
}_{k})-\phi(W_{i},\gamma_{0}^{\star},\alpha_{0}^{\star})-\bar{\phi}(\gamma
_{0}^{\star},\hat{\alpha}_{k})],\nonumber\\
\hat{R}_{2}  &  =\frac{1}{n_{k}}\sum_{i\in I_{k}}[\phi(W_{i},\hat{\gamma}%
_{k},\hat{\alpha}_{k})-\phi(W_{i},\hat{\gamma}_{k},\alpha_{0}^{\star}%
)-\phi(W_{i},\gamma_{0}^{\star},\hat{\alpha}_{k})+\phi(W_{i},\gamma_{0}^{\star
},\alpha_{0}^{\star})]\nonumber\\
&  =-\frac{1}{n_{k}}\sum_{i\in I_{k}}[\hat{\alpha}_{k}(X_{i})-\alpha_{0}%
^{\star}(X_{i})][\hat{\gamma}_{k}(X_{i})-\gamma_{0}^{\star}(X_{i})].
\end{align}
Define $\hat{\Delta}_{ik}=m(W_{i},\hat{\gamma}_{k}-\gamma_{0}^{\star})-\bar
{m}(\hat{\gamma}_{k}-\gamma_{0}^{\star})$ for $i\in I_{k}$ and let
$\mathcal{W}_{k}^{c}$ denote the observations $W_{i}$ for $i\notin I_{k}$.
Note that $\hat{\gamma}_{k}$ depends only on $\mathcal{W}_{k}^{c}$ by
construction. Then by independence of $\mathcal{W}_{k}^{c}$ and $\{W_{i},i\in
I_{k}\}$ we have $\Ep[\hat{\Delta}_{ik}|\mathcal{W}_{k}^{c}]=0.$ Also by
independence of the observations, $\Ep[\hat{\Delta}_{ik}\hat{\Delta}%
_{jk}|\mathcal{W}_{k}^{c}]=0$ for $i,j\in I_{k}.$ Furthermore, for $i\in
I_{k}$ $\Ep[\hat{\Delta}_{ik}^{2}|\mathcal{W}_{k}^{c}]\leq\int[m(w,\hat{\gamma
}_{k}-\gamma_{0}^{\star})]^{2}F_{0}(dw)$. Then by equation (\ref{L2cons}) we
have
\begin{align*}
\Ep\left[  \left(  \frac{1}{n_{k}}\sum_{i\in I_{k}}\hat{\Delta}_{ik}\right)
^{2}|\mathcal{W}_{k}^{c}\right]   &  =\frac{1}{n_{k}^{2}}\Ep\left[  \left(
\sum_{i\in I_{k}}\hat{\Delta}_{ik}\right)  |\mathcal{W}_{k}^{c}\right]
=\frac{1}{n_{k}^{2}}\sum_{i\in I_{k}}\Ep[\hat{\Delta}_{ik}^{2}|\mathcal{W}%
_{k}^{c}]\\
&  \leq\frac{1}{n_{k}}\int[m(w,\hat{\gamma}_{k}-\gamma_{0}^{\star})]^{2}%
F_{0}(dw)=o_{p}(\sigma^{2}/n_{k})=o_{p}(\sigma^{2}/n).
\end{align*}
The conditional Markov inequality then implies that $\sum_{i\in I_{k}}%
\hat{\Delta}_{ik}/n=o_{p}(\sigma/\sqrt{n}).$ The analogous results also hold
for $\hat{\Delta}_{ik}=\phi(W,\hat{\gamma}_{k},\alpha_{0}^{\star}%
)-\phi(W,\gamma_{0}^{\star},\alpha_{0}^{\star})-\bar{\phi}(\hat{\gamma}_{k},\alpha
_{0}^{\star})$ and $\hat{\Delta}_{ik}=\phi(W,\gamma_{0}^{\star},\hat{\alpha}%
_{k})-\phi(W,\gamma_{0}^{\star},\alpha_{0}^{\star})-\bar{\phi}(\gamma_{0}^{\star
},\hat{\alpha}_{k})$ by $\bar{\phi}(\gamma_{0}^{\star},\alpha_{0}^{\star})=0.$
Summing across the three terms in $\hat{R}_{1}$ gives $\hat{R}_{1}%
=o_{p}(\sigma/\sqrt{n})$.

Next let $\hat{\Delta}_{k}(x)=-[\hat{\alpha}_{k}(x)-\alpha_{0}^{\star}%
(x)][\hat{\gamma}_{k}(x)-\gamma_{0}^{\star}(x)].$ Then by the triangle and
Cauchy-Schwartz inequalities,%
\begin{align*}
\Ep[\left\vert R_{2}\right\vert |\mathcal{W}_{k}^{c}]  & \leq\int\left\vert
\hat{\Delta}_{k}(x)\right\vert F(dx)\leq\left\Vert \hat{\alpha}_{k}-\alpha
_{0}^{\star}\right\Vert _{P,2}\left\Vert \hat{\gamma}_{k}-\gamma_{0}^{\star
}\right\Vert _{P,2}=\sigma\sigma^{-1}\left\Vert \hat{\alpha}_{k}-\alpha
_{0}^{\star}\right\Vert _{P,2}\left\Vert \hat{\gamma}_{k}-\gamma_{0}^{\star
}\right\Vert _{P,2}\\
& \leq\sigma\sigma^{-1}(\left\Vert \hat{\alpha}_{k}-\alpha_{0}\right\Vert
_{P,2}+\left\Vert \alpha_{0}-\alpha_{0}^{\star}\right\Vert _{P,2})\left\Vert
\hat{\gamma}_{k}-\gamma_{0}^{\star}\right\Vert _{P,2}.
\end{align*}
By hypothesis $r_{2}^{\star}r_{1}^{\star}=o\left(  1/\sqrt{n}\right)  ,$ so
that by the conditional Markov inequality and the definition of $r_{2}^{\star
},$%
\[
\hat{R}_{2}=O_{p}(\sigma r_{2}^{\star}r_{1}^{\star})=o_{p}(\sigma/\sqrt{n}).
\]
The conclusion then follows by the triangle inequality. \qed

\section{Proofs for Section C}

\subsection{Proof of Lemma~\ref{lemma: key examples}} Use the same notation as in the proof of the previous lemma.
In all examples, $\alpha_ 0 \in L^2(F)$ and $\gamma \in L^2(F)$ imply that 
$| \langle \alpha_0, \gamma \rangle | < \| \alpha_0 \|_{P,2} \| \gamma\|_{P,2}< \infty$.
 
 Proof of claim (i).  In Example~\ref{ex:ate}, since $dF(x) =  \sum_{k=0}^1P[D=k|Z=z] 1(k=d)  dF(z)$ by the Bayes rule, we have
 $$ 
 \langle \alpha_0, \gamma \rangle = \int \gamma (d,z) \bar \ell(z)\frac{1(d=1) -1(d=0)}{P[D=d | Z=z]} dF(x) 
 $$$$
 = \Ep [ (\gamma (1,z) - \gamma(0,z)) \bar \ell(z)] 
 = \Ep [ (\gamma (1,z) - \gamma(0,z)) \ell(x)]  = \theta(\gamma),
 $$
using the law of iterated expectations to obtain the last line.  In Example~\ref{ex:policy1}, $\ell \alpha_ 0 \in L^2(F)$ means that the Radon--Nykodym derivatives $\frac{dF_1}{dF}$ and  $\frac{dF_0}{dF}$ exist on the support of $\ell$, so that
$$ 
\langle \alpha_0, \gamma \rangle  = \int \gamma  \ell \left(\frac{dF_1}{dF}- \frac{dF_0}{dF} \right) dF =
 \int \gamma \ell (dF_1- dF_0)  = \theta(\gamma).
 $$

We can demonstrate the claim for Example~\ref{ex:policy2} similarly to Example~\ref{ex:policy1}.

In Example~\ref{ex:deriv},
we can write
$$
 \langle \alpha_0  , \gamma \rangle  = - \int \int \gamma(x) \frac{\divg_d ( \ell(x) t(x) f(d|z))}{f(d|z)} f(d|z)\mathrm{d} d \mathrm{d}F(z)
$$$$
=  \int \int  \partial_d \gamma(x) ' t(x) \ell(x) f(d|z) \mathrm{d} d \mathrm{d} F(z) = \theta(\gamma),
$$
where we used the integration by parts and that  $\gamma(x)  \ell(x) t(x) f(d|z)$ vanishes on the boundary of $\mathcal{D}_z$.

The rest of the claim is immediate from Lemma~\ref{lemma: ERR}.  

Proof of claim (ii). We can refer to the case of linear regression discussed in Section~\ref{sec:linear}.

In what follows consider the case of $G>0$ and $\ell =1$.

In Example~\ref{ex:ate},  $M =  \Ep (b(1,Z) - b(0,Z))$. Suppose $P[D=0|Z] \in \{ 0,1 \}$  with probability in $[\pi, 1- \pi]$ for $\pi>0$, but such that $G>0$ (this puts restrictions on $b$). This is known as the case of failing overlap assumption in causal inference. Then $\alpha_0(X)$ is $\mathsf{na}$ with probability $\pi$.

In Example~\ref{ex:policy1} and~\ref{ex:policy2}, $M= \int b (d F_1- dF_0)$ is well defined, but $\alpha_0(X)= \mathsf{na}$ 
whenever $dF_1/dF$ and $dF_1/dF$ do not exist. For instance, $F_1$ and $F_0$ can have point masses, where $F$ does not, while retaining the same support as $F$.

In Example~\ref{ex:deriv},  take basis functions $b$ and a constant direction $t(X)=1$, such that $M = \Ep \partial_d b(D,Z)$ is well defined. Consider the case where $f(d|Z) = 0$ with positive probability so that $\alpha_0(X)= \mathsf{na}$ with this probability. \qed

\subsection{Proof of Lemma~\ref{lemma: additive}} The projection operator onto $\bar \Gamma_1 = L^2(F_1)$ is the conditional expectation with conditioning on $X_1$. The contractive property follows  from Jensen's inequality. \qed

\subsection{Proof of Lemma~\ref{lemma: gstr13}} The proof uses the fact that $m(W,\gamma)= m(X, \gamma)$, and
that
$$
\psi^\star(X)_0(W) =  -U_1 - \alpha_0^\star(X) U_2.
$$
Since $\Ep U_1 U_2 \alpha_0^\star(X) = 0$ by the LIE, using the bounded moments assumption we have:
$$
\sigma^2 = \Ep U_1^2 + \Ep U_2^2 \alpha_0^{\star 2} \geq \Ep [ \Ep(U_2^2 \mid X)  \alpha^{\star2}_0 (X)] 
\geq   \underline c^2 L^2.$$
The bound from above follows similarly:
$$
\sigma^2 = \Ep U_1^2 + \Ep U_2^2 \alpha_0^{\star 2} \leq \bar c^2 + \Ep [ \Ep(U_2^2 \mid X)  \alpha^{\star2}_0 (X)] 
\leq  \bar c^2 +  \bar c^2 L^2.
$$
Using the triangle inequality and bounded moments assumptions, we have:
\begin{eqnarray*}
\kappa \leq \| U_1\|_{P,3} + \| U_2 \alpha_0^\star\|_{P,3} & \leq &  \bar c + ( \Ep ( \Ep [ |U_2|^3\mid X] |\alpha^{\star }_0(X)|^3)   )^{1/3},\\
& \leq &  \bar c + \bar c \| \alpha^\star_0\|_{P,3} \leq \bar c ( 1+ c (L^2 \vee 1)),
\end{eqnarray*}
where the last line follows by assumption. \qed

\subsection{Proof of Lemma~\ref{lemma: lstr13}} We shall use that $m(W,\gamma) = m(X,\gamma)$,
and
$$
\psi_0^\star(W) =- U_1 - \alpha_0^\star(X) U_2.
$$
Then by $\Ep U_1 U_2 \alpha^\star_0(X) = 0$, holding by the LIE, we have
$$
\sigma^2 = \Ep U_1^2  + \Ep U_2^2 \alpha_0^{\star 2} =  \Ep U_1^2  +
\Ep ( \Ep [ U_2^2 \mid X ]  \alpha_0^{\star 2}(X) ).
$$

Then using the moment assumptions, we have
$$
 \underline c^2 \| \alpha_0^{\star}\|^2_{P,2}  \leq  \sigma^2 \leq \bar c^2 ( \| \ell\|^2_{P,2} +  \| \alpha_0^{\star}\|^2_{P,2}).
$$
Using the triangle inequality,
the LIE, and the bounded heteroscedasticity assumption, conclude
$$
\kappa \leq \| U_1\|_{P,3} + \| U_2 \alpha_0^\star\|_{P,3} \leq \bar c ( \| \ell\|_{P,3} + \| \alpha_0^\star \|_{P,3}).
$$   

For the case (a), $\alpha^\star_0(X) = \alpha_0(X;1)  \ell(X)$, using the assumed bound  $ \underline \alpha \leq \alpha_0(X;1) \leq \bar \alpha$ conclude that
$$
\underline \alpha \| \ell\|_{P,2} \leq L= \| \alpha^\star_0\|_{P,2} \leq \bar \alpha \| \ell\|_{P,2}, \quad 
 \| \alpha^\star_0\|_{P,3} \leq \bar \alpha \| \ell\|_{P,3}.
$$

For the case (b), $\alpha^\star_0(X_1) = \Ep [\alpha_0(X;1) \mid X_1]  \ell(X_1)$, so that by Jensen's inequality
$$\| \alpha^\star_0\|_{P,q} \leq \| \alpha_0(X;1)\ell(X_1) \|_{P,q} \leq  \bar \alpha \|\ell\|_{P,q} $$
and  using $$\underline  \alpha \leq \Ep [\alpha_0(X;1) \mid X_1],$$ holding because conditional expectation preserves
order, conclude that
$$
\| \alpha_0^{\star}\|^2_{P,2} = \Ep (\Ep [\alpha_0(X;1) \mid X_1]^2  \ell(X_1)^2 ) \geq \underline \alpha^2 \| \ell\|^2_{P,2}.
$$

Further, by change of variables in $\Bbb{R}^{p_1}$:
$u = (d_0 -d)/h, \text{ so that } \mathsf{d} u  = h^{-p_1} \mathsf{d} d,$
we have that
$$
 \| \ell\|^q_{P,q} \omega^q  = \int_{\Bbb{R}^{p_1}} h^{-p_1 q}|K^q ((d_0 -d)/h)| f_D(d) \mathsf{d} d  =  \int_{\Bbb{R}^{p_1}} h^{-p_1 (q-1) }|K^q (u)| f_D(d_0 - u h)  \mathsf{d} u 
$$
so that
$$
 h^{-p_1 (q - 1)/q}  \underline f^{1/q}  \left(\int |K|^q\right)^{1/q} \leq \| \ell\|_{P,q} \omega \leq  h^{-p_1 (q - 1)/q}  \bar f^{1/q}  \left(\tiny{\int} |K|^q\right)^{1/q}.
$$ 
Further, we have that
$$
\omega  = \int  h^{-p_1} K((d_0-d)/h) f_D(d) \mathsf{d} d = \int   K(u) f_D(d_0- u h) \mathsf{d} u.
$$
Using the Taylor expansion in $h$ around $h=0$ and the Holder inequality:
$$
|\omega - f_D(d_0)| =  \left|   \int   K(u) h \partial f_D(d_0- u \tilde h)' u \mathsf{d} u \right |  \leq   h \bar f' \int \|u\|_\infty | K(u)| du,
$$
for some $0 \leq \tilde h  \leq h$. Hence for all $h< h_1 < h_0$,  with $h_1$ depending only on $(K, \bar f', \underline{f}, \bar f)$:
$$ \underline{f}/2  \leq \omega \leq 2 \bar f.$$

In summary, we have the following non-asymptotic bounds for all $0 < h < h_1$:
$$
\underline{c}  \underline \alpha \| \ell\|_{P,2}  \leq \sigma \leq \bar c \sqrt{1+ \bar \alpha} \| \ell\|_{P,2}, \quad  \underline{ \alpha} \| \ell\|_{P,2}  \leq L \leq \bar \alpha \| \ell\|_{P,2},  \quad
\kappa \leq  \bar c (1 + \bar \alpha ) \| \ell \|_{P,3},
$$
where
$$
 h^{-p_1 (q - 1)/q}  \underline f^{1/q}  \left(\int |K|^q\right)^{1/q} /(2 \bar f) \leq \| \ell\|_{P,q}  \leq  h^{-p_1 (q - 1)/q}  \bar f^{1/q}  \left(\tiny{\int} |K|^q\right)^{1/q}  2/\underline{f}.
$$ 

As $h \to 0$, we have that
$$
\sigma \asymp L \asymp \| \ell\|_{P,2} \asymp h^{-p_1/2},   \ \  \kappa \lesssim h^{-2p_1/3},  \ \ \kappa/\sigma \lesssim h^{-p_1/6}.
$$

\qed

\subsection{Proof of Lemma~\ref{lemma: lstr4}} Similarly to the proof of Lemma~\ref{lemma: lstr13}, using the LIE
and bounded heteroscedasticity, we obtain
$$
 \| \alpha^\star_0 \|_{P,2}^2 \underline c^2 
\leq \sigma^2 \leq \| \ell\|_{P,2}^2 \bar c^2 + \| \alpha^\star_0 \|_{P,2}^2 \bar c^2,
$$and by the triangle inequality $$
\kappa  \leq \| \ell\|_{P,3} \bar c + \| \alpha^\star_0 \|_{P,3} \bar c.
$$

It remains to bound $\| \alpha^\star_0\|_{P,q}$. To help this, introduce notation
$$ v(X) := f(D\mid Z).$$

Case (a). We have that
$$
\alpha^\star_0 = \alpha_0 = \divg_d (\ell) t +   \divg_d (t) \ell + \divg_d(v) \ell t/v.
$$
By the triangle inequality,
$$
\| \alpha^\star_0 \|_{P,q} \leq \| \divg_d (\ell) t\|_{P,q} + \|  \divg_d(t) \ell  \|_{P,q} +  \|\divg_d(v) \ell t/v\|_{P,q},
$$$$
\| \alpha^\star_0 \|_{P,2} \geq \| \divg_d (\ell) t\|_{P,2} -  \|  \divg_d(t) \ell  \|_{P,2} -  \|\divg_d (v) \ell t/v\|_{P,2}.
$$
Using the bounds assumed in the Lemma, we have
\begin{eqnarray*}
\| \divg_d (\ell) t\|_{P,q} \leq \| \divg_d (\ell) \|_{P,q} \bar t;  \quad  \|  \divg_d(t) \ell  \|_{P,q} \leq \bar t'  \|\ell  \|_{P,q};
\quad   \|\divg_d(v) \ell t/v\|_{P,q} \leq \| \ell \|_{P,q} (\bar f' \bar t/\underline f).
\end{eqnarray*}  
By the  proof of Lemma~\ref{lemma: lstr13}, for all $h< h_1 < h_0$,  with $h_1$ depending only on $(K, \bar f', \underline{f}, \bar f)$:
$$ \underline{f}/2  \leq \omega \leq 2 \bar f,$$ and
$$
 h^{-p_1 (q - 1)/q}  \underline f^{1/q}  \left(\int |K|^q\right)^{1/q} /(2 \bar f) \leq \| \ell\|_{P,q}  \leq  h^{-p_1 (q - 1)/q}  \bar f^{1/q}  \left(\tiny{\int} |K|^q\right)^{1/q}  2/\underline{f}.
$$  

Furthermore, by the LIE and the assumed lower bounds in the statement:
\begin{eqnarray*}
\| \divg_d (\ell) t\|_{P,2}^2 & =& \Ep [\divg(\ell)^2 \Ep (t^2 | D )] \\
 & = &   \omega^{-2} h^{-2} h^{- p_12}  \int (\divg K((d_0 -d)/h  )^2  \Ep (t^2 | D=d ) f(d) \mathsf{d} d\\
 & = &  \omega^{-2} h^{-2} h^{- p_1} \int (\divg K(u))^2\Ep (t^2 | D=d_0 - hu )f(d_0 - hu) du \\
 & \geq &  (2 \bar f )^{-2} h^{-2} h^{-p_1} \underline t^2 \underline f   \int (\divg K)^2,
\end{eqnarray*} 
and similarly
$$
\| \divg_d (\ell) \|_{P,q}^q  \leq  \omega^{-q} h^{-q} h^{-p_1(q-1)} \bar f   \int |\divg K|^q
\leq (\underline f/2) ^{-q} h^{-q} h^{-p_1(q-1)} \bar f   \int |\divg K|^q
$$

Case (b). Here we have, using the notation as above
\begin{eqnarray*}
\alpha^\star_0(X_1)  =   \Ep[\alpha_0 \mid X_1] & = & \divg_d (\ell (X_1) ) \Ep[t(X_1) \mid X_1] \\
& + &    
\Ep[\divg_d (t(X) \mid X_1] \ell(X_1) +  \Ep [\divg_d(v(X))  t(X)/v(X) \mid X_1] \ell(X_1).
\end{eqnarray*}
Then by contractive property of the conditional expectation
$\| \alpha^\star_0 \|_{P,q} \leq \| \alpha_0 \|_{P,q}$, so the upper bounds
apply from case (a).

We only need to establish
lower bound on $\| \alpha^\star_0 \|_{P,2}$. By the triangle inequality,
$$
\| \alpha^\star \|_{P,2} \geq \| \divg_d (\ell) \Ep [t \mid X_1] \|_{P,2} -  \| \Ep[\divg_d (t) \mid X_1] \ell  \|_{P,2} -  \|\Ep[\divg_d (t) \mid X_1] \ell \|_{P,2}.
$$
By Jensen's inequality, and using the same calculations as in case (a):
$$
 \|  \divg_d (\ell (X_1) ) \Ep[t(X_1) \mid X_1] \|_{P,2} \leq  \|  \divg_d (\ell (X_1) )t(X_1) \|_{P,2} \leq  \bar t \| \divg_d (\ell) \|_{P,q};
$$ $$
 \| \Ep[\divg_d (t) \mid X_1] \ell  \|_{P,2} \leq \|  \divg_d (t) \ell  \|_{P,q} \leq \bar t'  \|\ell  \|_{P,q};
 $$ $$
 \|\Ep[\divg_d(v)  t/v \mid X_1] \ell \|_{P,2} \leq \|\divg_d(v) \ell t/v\|_{P,q} \leq \| \ell \|_{P,q} (\bar f' \bar t/\underline f).
$$

And, similarly to the calculation above
\begin{eqnarray*}
\|  \divg_d (\ell) \Ep [t \mid X_1]\|_{P,2}^2 & =& \Ep [\divg_d(\ell)^2 \Ep ( (\Ep[t \mid X_1])^2 | D )] \\
 & = &  \omega^{-2} h^{-2} h^{- p_12}  \int (\divg K((d_0 -d)/h  )^2  \Ep ( (\Ep[t \mid X_1])^2 | D=d) f(d) \mathsf{d} d\\
 & = & \omega^{-2}  h^{-2} h^{- p_1} \int (\divg K(u)^2\Ep ((\Ep[t \mid X_1])^2 | D=d_0 - hu )f(d_0 - hu) du \\
 & \geq & \omega^{-2}  h^{-2} h^{-p_1} \underline t^2 \underline f   \int (\divg K)^2 \\
 &   \geq &   (2 \bar f )^{-2} h^{-2} h^{-p_1} \underline t^2 \underline f   \int (\divg K)^2,
\end{eqnarray*} 
using the assumed bound $\Ep ((\Ep[t \mid X_1])^2 | D=d) \geq \underline t^2$ for $d \in N_h(d_0)$.

In either case (a) or (b), we now summarize the bounds asymptotically by letting $h \searrow 0$:
$$
 L \lesssim \sigma  \lesssim h^{-p_1/2} (1+ h^{-1}),   \quad h^{-p_1/2}  (h^{-1} - 1) \lesssim L \lesssim h^{-p_1/2}  (h^{-1} + 1),$$
$$\kappa \lesssim h^{-2p_1/3}  (h^{-1}+1), \quad \kappa/\sigma \lesssim h^{-p_1/6}.
$$ \qed

\subsection{Proof of Lemma~\ref{lemma: bias}}  Introduce $m(d): = \Ep [m(W, \gamma_0^\star) \mid D= d] $
and note
$$
\vartheta_1(h) =  \int m(d) h^{-p_1} K((d_0-d)/h) f_D(d) \mathsf{d} d= \int m(d_0 - h u) K(u) f_D(d_0 - hu)\mathsf{d} u,
$$
$$
\vartheta_2(h) =  \int  h^{-p_1} K((d_0-d)/h) f_D(d) \mathsf{d} d = \int   K(u) f_D(d_0- u h) \mathsf{d} u.
$$
Note that by $\int K  =1 $,
$$
\vartheta_1(0) = m(d_0) f_D(d_0), \quad \vartheta_2(0) = f_D(d_0). 
$$
Hence$$
\theta(\gamma_0^\star; \ell_h) = \frac{\vartheta_1(h) }{\vartheta_2(h)},  \quad \theta(\gamma_0^\star; \ell_0) := \frac{\vartheta_1(0) }{\vartheta_2(0)} = m(d_0). $$
By the standard argument to control the bias of the higher-order kernel smoothers, e.g. by Lemma~B2 in \cite{newey1994kernel}, which employs the Taylor expansion of order $\mathsf{v}$ in $h$ around $h=0$, for some constants $A_{\mathsf{v}}$ that depend only on  $\mathsf{v}$: 
$$
| \vartheta_1(h) - \vartheta_1(0)| \leq A_{\mathsf{v}} h^{ \mathsf{v}} \bar g_\mathsf{v} \int \|  u\|^{\mathsf{v}} | K(u)| du,
$$$$
| \vartheta_2(h) - \vartheta_2(0)| \leq A_{\mathsf{v}} h^{\mathsf{v}} \bar f_\mathsf{v} \int \|  u\|^{\mathsf{v}} | K(u)| du,
$$
where $\mathsf{v} = \mathsf{o} \wedge {\mathrm{sm}}$. Then using the relation
\begin{eqnarray*}
\frac{\vartheta_1(h) }{\vartheta_2(h)} - \frac{\vartheta_1(0) }{\vartheta_2(0)}  = \left (\begin{array}{l}
 \vartheta^{-1}_2(0) (\vartheta_1(h)  - \vartheta_1(0)) +  \vartheta_1(0) (\vartheta_2^{-1}(h)-\vartheta_2^{-1}(0))\\
  +   (\vartheta_1(h)  - \vartheta_1(0))(\vartheta_2^{-1}(h)-\vartheta_2^{-1}(0)) 
 \end{array}\right), \end{eqnarray*}
we deduce the following bound that applies for all $h< h_1\leq h_0$,
$$
|\theta(\gamma_0^\star; \ell_h)  - \theta(\gamma_0^\star; \ell_0)| \leq  \left | \frac{\vartheta_1(h) }{\vartheta_2(h)} - \frac{\vartheta_1(0) }{\vartheta_2(0)}  \right | \leq C h^{\mathsf{v}}, 
$$
where the constant $C$ and $h_1$ depend  only on $K, \mathsf{v}, \bar g_{\mathsf{v}}$,  $\bar f_{\mathsf{v}}$, $\underline f$.  \qed

\end{document}